\documentclass[10pt,twocolumn,letterpaper]{article}

\usepackage[pagenumbers]{cvpr}

\usepackage{graphicx}
\usepackage{booktabs}
\usepackage{dblfloatfix}

\usepackage[accsupp]{axessibility}  

\usepackage{xspace}
\usepackage[dvipsnames]{xcolor}
\usepackage{comment}
\usepackage{adjustbox}
\usepackage{array}
\usepackage{makecell}
\usepackage{pifont}
\usepackage{multirow}
\usepackage{flushend}
\usepackage{colortbl}
\usepackage{enumitem}
\usepackage{comment}
\usepackage{sidecap}
\newcolumntype{R}[2]{%
    >{\adjustbox{angle=#1,lap=1.3\width-(#2)}\bgroup}%
    l%
    <{\egroup}%
}
\usepackage{hyperref}
\usepackage[]{pgfplots,pgfplotstable}
\usepackage{algorithm}
\usepackage{algpseudocode}

\usepackage[]{tikz}
\usetikzlibrary{positioning, math, fit, backgrounds}

\usepackage{orcidlink}

\makeatletter
\renewcommand\paragraph{\@startsection{paragraph}{4}{\z@}%
    {-6\p@ \@plus -3\p@ \@minus -3\p@}%
    {-0.5em \@plus -0.22em \@minus -0.1em}%
    {\normalfont\normalsize\itshape}}
\makeatother

\definecolor{myblue}{rgb}{0.19, 0.55, 0.91}
\definecolor{myred}{rgb}{0.82, 0.1, 0.26}
\newcommand{\cmark}{\textcolor{myblue}{\ding{51}}}
\newcommand{\xmark}{\textcolor{myred}{\ding{55}}}
\newcommand{\ra}[1]{\renewcommand{\arraystretch}{#1}}

\def\ours{MILAN\xspace} 

\title{MILAN: Milli-Annotations for Lidar Semantic Segmentation}

\author{Nermin Samet\orcidlink{0000-0001-9247-2504}$^1$ \qquad
Gilles Puy\orcidlink{0000-0003-3502-980X}$^1$ \qquad
Oriane Siméoni\orcidlink{0000-0003-3232-8978}$^1$ \qquad
Renaud Marlet\orcidlink{0000-0003-1612-1758}$^{1,2}$ \\ 
\hspace{-5mm}\textsuperscript{1}Valeo.ai, Paris, France
\hspace{2.5mm}\textsuperscript{2}LIGM, Ecole des Ponts, Univ Gustave Eiffel, CNRS, Marne-la-Vall\'ee, France
}

\begin{document}


\maketitle

\begin{abstract}
    Annotating lidar point clouds for autonomous driving is a notoriously expensive and time-consuming task. In this work, we show that the quality of recent self-supervised lidar scan representations allows a great reduction of the annotation cost. Our method has two main steps. First, we show that self-supervised representations allow a simple and direct selection of highly informative lidar scans to annotate: training a network on these selected scans leads to much better results than a random selection of scans and, more interestingly, to results on par with selections made by SOTA active learning methods. In a second step, we leverage the same self-supervised representations to cluster points in our selected scans. Asking the annotator to classify each cluster, with a single click per cluster, then permits us to close the gap with fully-annotated training sets, while only requiring one thousandth of the point labels.
\end{abstract}

\section{Introduction}
\label{sec:intro}

Despite advances in open vocabulary models for lidar data \cite{ding2023pla, chen2023clip2scene, peng2023openscene}, the performance of these models for semantic segmentation still lags behind the performance of models trained specifically with manually annotated data. Several research directions have been explored to reduced the cost of these manual annotations while targeting a similar performance as with full supervision.

First, observing that datasets often contain useless samples (redundant or too easy), active learning methods iteratively select best subsets of data to annotate, trying to maximize the performance when training using the labeled part of the dataset. The performance of these methods however varies significantly with the very first subset of data to annotate, which is generally a random seed, i.e., a randomly selected subset of scans. SeedAL~\cite{samet2023seedal} was recently proposed to systematically select a good first set of scans to label. However, even with such a warm start, the gap with full supervision remains hard to close, even with (typically) 5\% of labeled points \cite{wu2021redal, hu2022lidal}.

Second, self-supervised methods leverage a pretext tasks and a large set of unlabeled data to pretrain a backbone. The pretrained backbone can then be finetuned on a small amount of annotated, data achieving a higher performance than the same backbone trained from scratch on the same data \cite{slidr, mahmoud2023stslidr, puy2024scalr}. Even in the best current setting~\cite{puy2024scalr}, the gap with full supervision remains larger than 10\,\%pt of mIoU when finetuning on 1\% of labeled data.

Third, semi-supervised methods leverage both a small set of labeled data and a possibly large set unlabeled data to train a model. These methods use various techniques to extend annotations to unlabeled data and learn from that, boosting the overall performance of the trained model \cite{lee2013pseudo, jiang2021guided, hou2021contrastsc, hu2022sqn, liu2022ws3d, liu2023cpcm, kong2023lasermix}. While some of them are close to bridging the gap with full supervision \cite{li2022hybridcr}, they often require 1\% of labeled data to do so. A few of them operate with as low as 0.1\% of labeled data, while getting a performance close to full supervision \cite{li2022coarse3d, hu2022sqn, xie2023annotator}, however with backbones and settings that are not state-of-the-art (SOTA).

Finally, a method such as LESS \cite{liu2022less} proposes to facilitate the annotation process by asking an annotator to label pre-computed segments rather than points. This process is fast as only one or a few clicks are required to annotate a segment. Semi-supervised techniques and a multi-scan treatment are further used to make the most of the partially pre-annotated dataset. On a dataset like SemanticKITTI~\cite{semantickitti}, which is however known to display little diversity, LESS shows a very impressive performance: it bridges the gap to full supervision with as few as 0.1\% of manually labeled points, and it stays only 5\,\%pt of mIoU below full supervision at 0.01\% of labeling. Still, the method is complex and the task of the annotator actually is significantly harder, as explained below.

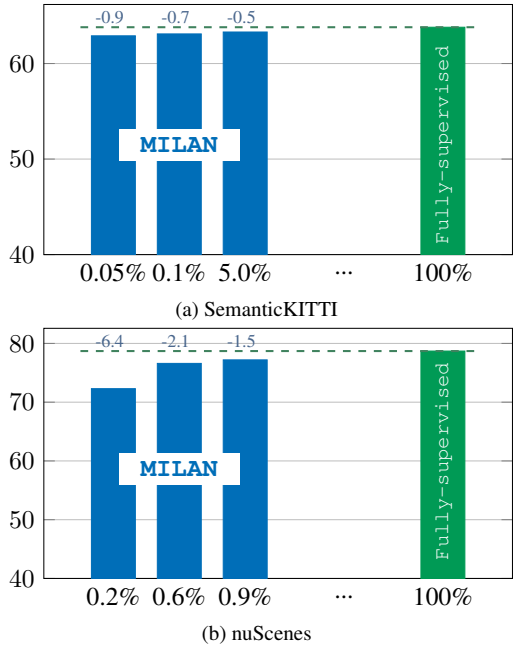
\begin{figure}[t!]
    \centering
    \begin{subfigure}[c]{0.4\textwidth}
        \centering
        \resizebox{\linewidth}{!}{
           \begin{tikzpicture}
            \begin{axis}[
                ymin=40,
                width=6cm,
                height=5cm,
                x=0.9cm, 
                enlarge x limits={abs=0.5cm}, 
                xtick={1,2,3,6},
                x tick style={draw=none},
                xticklabels={0.05\%, 0.1\%, 5.0\%, 100\%},
                ymajorgrids = true,
                every axis plot/.append style={
                  ybar,
                  bar width=.6cm,
                  bar shift=0pt,
                  fill
                },
                legend cell align=left,
                legend style={
                        at={(1,1.05)},
                        anchor=south east,
                        column sep=1ex
                },
              ]
                \addplot[NavyBlue] coordinates {
                    (1,   62.9) 
                    (2,   63.1) 
                    (3,   63.3) 
                };
                \addplot[ForestGreen] coordinates {
                    (6,  63.8) 
                };
                
                \addplot[style={
                    solid,
                    thick,
                    mark=none,
                    mark options={solid},
                    smooth, 
                    Green!60!black,
                    dashed,
                }]
                coordinates {(0.5, 63.8) (6.5,63.8)};
                \coordinate (A) at (4.95cm,1.5cm);
                \coordinate (B) at (1.35cm,1.5cm);
                \coordinate (C) at (3.6cm,-0.25cm);
                \coordinate (first) at (0.4cm,3.25cm);
                \coordinate (second) at (1.3cm,3.25cm);
                \coordinate (third) at (2.2cm,3.25cm);
                \coordinate (fourth) at (2.8cm,3.25cm);
            \end{axis}
            \node[rotate=90] at (A) {\textcolor{white}{\texttt{\footnotesize {Fully-supervised}}}};
            \node[fill=white] at (B) {\textcolor{NavyBlue}{\bf ~~\texttt{\ours}~~}};
            \node at (C) {...};
            \node[] at (first) {\textcolor{NavyBlue!60!black}{\scriptsize{-0.9}}};
            \node[] at (second) {\textcolor{NavyBlue!60!black}{\scriptsize{-0.7}}};
            \node[] at (third) {\textcolor{NavyBlue!60!black}{\scriptsize{-0.5}}};
        \end{tikzpicture}   
        }
        \caption{SemanticKITTI}
    \end{subfigure}
    \begin{subfigure}[c]{0.4\textwidth}
        \centering
        \resizebox{\linewidth}{!}{
           \begin{tikzpicture}
            \begin{axis}[
                ymin=40,
                width=6cm,
                height=5cm,
                x=0.9cm, 
                enlarge x limits={abs=0.5cm}, 
                xtick={1,2,3,6},
                x tick style={draw=none},
                xticklabels={0.2\%, 0.6\%, 0.9\%, 100\%},
                ymajorgrids = true,
                every axis plot/.append style={
                  ybar,
                  bar width=.6cm,
                  bar shift=0pt,
                  fill
                },
                legend cell align=left,
                legend style={
                        at={(1,1.05)},
                        anchor=south east,
                        column sep=1ex
                },
              ]
                \addplot[NavyBlue] coordinates {
                    (1,   72.3) 
                    (2,   76.6) 
                    (3,   77.2) 
                };
                \addplot[ForestGreen] coordinates {
                    (6,  78.7) 
                };
                
                \addplot[style={
                    solid,
                    thick,
                    mark=none,
                    mark options={solid},
                    smooth, 
                    Green!60!black,
                    dashed,
                }]
                coordinates {(0.5, 78.7) (6.5,78.7)};
                \coordinate (A) at (4.95cm,1.5cm);
                \coordinate (B) at (1.35cm,1.5cm);
                \coordinate (C) at (3.6cm,-0.25cm);
                \coordinate (first) at (0.4cm,3.25cm);
                \coordinate (second) at (1.3cm,3.25cm);
                \coordinate (third) at (2.2cm,3.25cm);
                \coordinate (fourth) at (2.8cm,3.25cm);
                \coordinate (asd) at (2.8cm,3.5cm);

            \end{axis}
            \node[rotate=90] at (A) {\textcolor{white}{\texttt{ \footnotesize{Fully-supervised}}}};
            \node[fill=white] at (B) {\textcolor{NavyBlue}{\bf ~~\texttt{\ours}~~}};
            \node at (C) {...};
            \node[] at (first) {\textcolor{NavyBlue!60!black}{\scriptsize{-6.4}}};
            \node[] at (second) {\textcolor{NavyBlue!60!black}{\scriptsize{-2.1}}};
            \node[] at (third) {\textcolor{NavyBlue!60!black}{\scriptsize{-1.5}}};
        \end{tikzpicture}
        }
        \caption{nuScenes}
    \end{subfigure}
    
    \caption{Performance (mIoU\%) obtained with \ours using WaffleIron \cite{puy23waffleiron} on SemanticKITTI and nuScenes for different (very) low levels of annotation. We compare our results to those of a model trained in a fully-supervised fashion with $100\%$ of the labels. We observe that with just $0.05\%$ of annotated points on SemanticKITTI, we obtain results on-par with those of the fully-supervised model. On the more challenging nuScenes dataset, we reach results similar to the fully supervised baseline with as little as 0.9\% of annotated points. 
    }
    \label{fig:teaser}
\end{figure}

In this work, we propose a simple method that targets a three-order-of-magnitude reduction of the manually-labeled points while reaching a performance close to a fully-annotated training set (see \cref{fig:teaser}). 

\begin{figure*}[t!]
\centering
\includegraphics[width=0.98\linewidth]{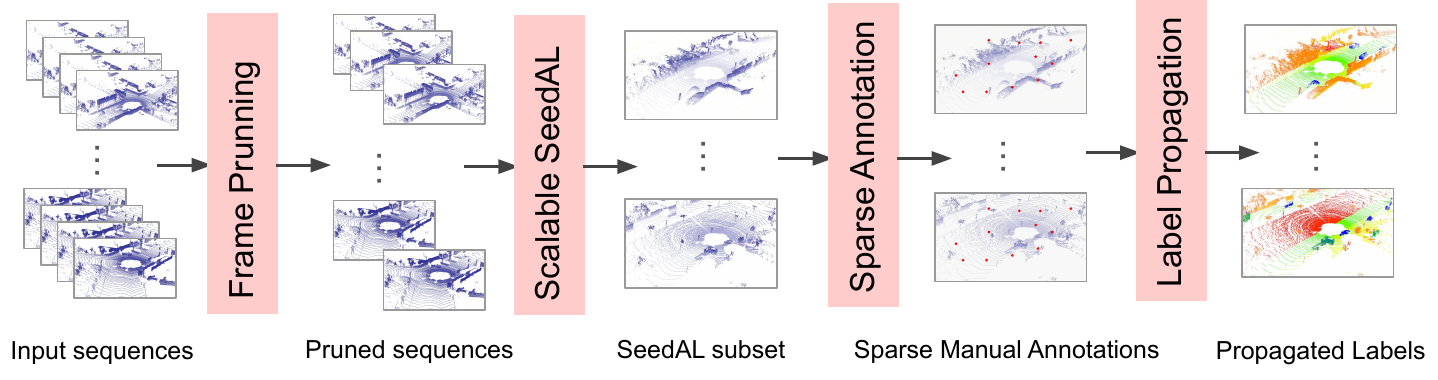}
\caption{\textbf{\ours's annotation pipeline}. The sequences of scans are first pruned to remove highly similar consecutive frames. Then a subset of scans with high diversity are selected using our scalable SeedAL method. For each selected scan, the points are clustered using self-supervised features. One point in each cluster is presented to an annotator, to prove a semantic label. The resulting sparse annotations are then propagated to the whole scan by giving the same label to all points falling in the same cluster. This scarce-annotation pipeline is followed by self-training.}
\label{fig:milan}
\end{figure*}

Our method, dubbed \ours for milli-annotation, starts with a smart selection of the most relevant scans, improving over SeedAL~\cite{samet2023seedal}. It then constructs point clusters that have high chances to contain points of a single class. Both the frame selection and the clustering build upon strong self-supervised lidar features \cite{puy2024scalr}. Next, the annotator is asked to label each cluster center, and the label is propagated to the whole cluster. This pipeline is illustrated in \cref{fig:milan}. Last, a model is trained with a teacher-student semi-supervised approach, both to extend labels to unselected frames (the rest of the dataset) and to remove some noise in impure clusters, containing a few points actually belonging to another class. By labeling only cluster centers in selected frames, it is possible to annotate only on the order of one thousandth of the points, while label propagation and semi-supervision yield a model with a performance close to a fully-supervised network. 
Our contributions are as follows: 
\begin{itemize}[topsep=2pt, itemsep=2pt]
    \item[\checkmark] Our selection of scans to annotate scales better than SOTA selection, allowing the direct selection of a large pool of relevant scans to mine from. As we can directly select a given number of scans, we save multiple active-learning retraining on various amount of annotated data. Besides, when fully annotated, training on our direct selections has a similar or better performance than training on iterative selections made by SOTA active learning methods.
    
    \item[\checkmark] We show that the quality of SOTA self-supervised features is enough to propagate information from as low as $1\%$ of annotated to points per scan to full scans with more than $87$\% of average classwise accuracy.
    
    \item[\checkmark] Semi-supervised training from our $0.1\%$ and $0.9\%$ of annotated points permits us to reach $98.9\%$ and $98.0\%$ of the performance one would reach with a fully-annotated training set of SemanticKITTI and nuScenes, respectively. 
     
\end{itemize}

\section{Related work}

\paragraph{Few-shot finetuning.}

A number of general models providing good 3D features for downstream tasks have been developed, either using self-supervision at 3D level \cite{boulch2023also, tarl, wu2023spatiotemporal, min2023occupancy, sautier2024bevcontrast} or distilling high-quality 2D features from associated images into 3D \cite{slidr, mahmoud2023stslidr, liu2023seal, puy2024scalr}. However, these approaches are generally evaluated on linear probing or after finetuning with full-supervision on a small dataset fraction (typically 1\% or 10\%). On such fractions, pretraining leads to significant improvement compared to fully-supervised training from scratch. However, as the remaining unlabeled fraction of the dataset is ignored, the performance gap compared to training with full supervision on the whole dataset remains extremely wide, typically on the order of -30 mIoU \%pt, with 1\% of points labeled~\cite{slidr, liu2023seal}.

\paragraph{Active learning (AL).}

Starting from a small initial annotated subset, active learning methods select, based on these few labels, a new small unlabeled subset to be annotated, that has high chances to lead to a good performance improvement when training a model with these additional labels \cite{coreset, lin2020segent, wu2021redal, hu2022lidal, xie2023annotator}. This process is repeated, with model retraining, until a given, accumulated annotation budget is spent, often expressed as a total number of manually labeled points. In contrast, we select data in one shot, without iterated retraining.

\paragraph{Annotation granularity.}

Rather than selecting entire frames to be fully annotated, recent methods are finer-grained, selecting regions~\cite{wu2021redal, hu2022lidal}, e.g., based on supervoxels \cite{papon2013vccs}, possibly over a frame accumulation to factorize annotations~\cite{unal2023dial}. In contrast, rather than fully annotating whole scans or regions, we only need a few points to be annotated.

Some methods leverage just one or a few labeled points too, often by propagating the label to a region, e.g., a superpoint~\cite{sspcnet}, a supervoxel~\cite{ye2022supervoxel}, or a cluster based on point distance~\cite{liu2022less}. Such geometry-based regions work well for dense and uniform point clouds, as often available for indoor scenes, but can be unreliable for sparse point clouds of non-uniform densities, as available with outdoor lidar scans. In contrast, we exploit more robust clusters based on point features.

One way to significantly further reduce the annotations is operate a form of label propagation across frames \cite{shi2022weakly, unal2023dial}

\paragraph{Annotator task.}

What the annotator is asked to do may conceal subtle but significant differences. LESS~\cite{liu2022less} displays a geometric cluster of point, and then asks the annotator to identify possible different classes in the cluster and click on one arbitrary point per class. OTOC~\cite{liu2021otoc} does not present any explicit region but rather asks the annotator to explore the whole scene, identifying all objects and clicking one arbitrary point per object in the frame. In contrast, we do not display regions nor ask the annotator to explore a region or the full scene; we only iteratively highlight a well-chosen point and ask the annotator for its class. It takes less time as only the class has to be chosen, with a single keystroke or click, rather than two for choosing both a point and a class. More importantly, it represents a much lower cognitive load as the annotator is not required to fully explore explicit~\cite{liu2022less} or implicit~\cite{liu2021otoc} regions; it is thus faster and less error-prone.

\paragraph{Data selection.}

Leveraging a model already trained on some data with known labels, a common selection strategy is to find a subset of unlabeled data on which the uncertainty of the model is maximum~\cite{choi2021active, unal2023dial}, hoping that labeling and training on these additional data yield better predictions. Another strategy is to maximize diversity, e.g., at gradient level in a batch \cite{Ash2020Deep} or regarding underlying features \cite{sinha20219variational}, to rapidly cover variety in the dataset and expose the model to a wide range of samples. The size of geometric clusters of points may also be used favor diversity~\cite{wei2023basal}. Our method focuses on diversity maximization, which is evaluated without any new model training, globally and efficiently.

A number of diversity-based methods exploit features made from the pool of labeled data \cite{sinha20219variational, sspcnet, buchert2022exploiting}. Instead, we exploit high-quality features of a pretrained model~\cite{puy2024scalr}. So does SeedAL~\cite{samet2023seedal}, that maximizes a feature-based diversity measure to create a first good subset for active learning methods. However, SeedAL only selects entire scenes rather than regions or points, and it does not scale well to a large pool of data because of the quadratic size of the underlying linear program. Instead, we propose a scalable variant of SeedAL, which we use to select a small number of relevant frames, in which we identify points to annotate.

\paragraph{Semi-supervised learning.}

After a subset of data is selected and annotated, possibly with some form of partial label propagation, the question is how to make the most of unlabeled data for the final training. Various general techniques have been proposed for this semi-supervised learning, which can be combined with priors specific to point clouds and in particular to lidar scans, including augmentations \cite{li2022hybridcr, liu2023cpcm}.

A popular approach is based on pseudo-labeling, where unannotated data receive (pseudo-)labels from the currently trained network and the training loss balances the significance of true labels and pseudo-labels, possibly taking into account the confidence in estimating pseudo-labels \cite{lee2013pseudo}.

Some methods leverage, implicitly or explicitly, a form of self-supervised learning at loss level \cite{jiang2021guided, yang2022mil, hu2022sqn, liu2023cpcm}. In contrast, we leverage self-supervised features from a pretrained model. Although these features are not necessarily specialized for the data considered, their quality is often higher as such models are pretrained on a much larger quantity of data.

Another common technique to train with pseudo-labels on unlabeled data is to do self-training or self-distillation, typically with a teacher-student architecture involving an exponential moving average (EMA). The architecture and loss may differ from one method to another \cite{unal2023dial, kong2023lasermix, liu2023weak}. Pseudo-labels can also be combined with augmentations concerning multiple frames, mixing labeled and unlabeled scans \cite{kong2023lasermix}. Alternatively, WS3D \cite{liu2022ws3d} implement a self-supervised objective based on point cloud geometry.

Other approaches include prototype learning \cite{li2022coarse3d, li2022hybridcr, su2023weakly} and central-point-based attention \cite{tran2024pointct}. Images may also be used in a multi-modal setting~\cite{chen2023clip2scene}.

\section{Method}
\label{sec:method}

Our method consists of three stages. The first stage (\cref{sec:frameselection}) selects a fraction of representative frames of the dataset based on features from a pre-trained model. The second stage (\cref{sec:manualannotation}) selects a fraction of points in these frames to be manually annotated based on feature clustering. For each cluster, one point is annotated and this label is propagated to all points in the cluster. The third stage (\cref{sec:teacherstudent}) consists in training a network using the pseudo-labeled scans and leveraging a teacher-student technique to also exploit the unannotated scans.

\subsection{Frame selection}
\label{sec:frameselection}

The selection of frames, i.e., scans, is performed in two steps. First, as there is a lot of redundancy between consecutive frames, we prune each sequence as follows. We consider the first frame as the base frame and drop all successive frames until one differs enough from the base frame in terms of frame feature, using cosine similarity with a fixed threshold~$\tau$. The differing frame is then considered as the new base frame and the process is repeated until the end of the sequence. This pruning strategy is similar to the pruning used in SeedAL \cite[Sec.\,A]{samet2023seedal}. We however use only 3D features from ScaLR \cite{puy2024scalr} (distilled from DINOv2 \cite{oquab2023dinov2}) while SeedAL relies on features of images associated to scans (obtained from DINOv1 \cite{caron2021dino}). In fact, our method operates only on scans; it does not need images.

After sequence pruning, to choose the most diverse frames among the remaining frames, we use a scalable variant of SeedAL, which we design. The fact is that the linear program used in SeedAL to maximize diversity, cannot scale to a very large number of frames (not more than hundreds). Concretely, SeedAL tries to find the largest subset of frames, within a maximum size, while maximizing the sum of the pairwise frame dissimilarity, i.e., SeedAL's combined intra and inter-scene diversity measure~$e_{ij}$ \cite[Sec.\,4.1]{samet2023seedal}. The bottleneck of this optimization however is not the quadratic evaluation of all pairwise similarities but the maximization of a subset of point clouds within a maximum constraint of the total size. Indeed, it relies on mixed integer programming with a number of variables that is also quadratic in the number frames.

We simplify the problem by assigning a diversity score to each frame, which allows us to sort frames and keep only the most diverse ones within a size limit. The diversity score of a frame is defined as the average dissimilarity w.r.t.\ all other frames. This much simpler process is only quadratic in the number of frames, which allows to easily scale, e.g., to tens of thousands of frames, which is totally unreachable by SeedAL because of the underlying optimization process. While it does not guarantee to find an optimum subset, it generally finds a very good one.

\subsection{Manual annotation and pseudo-labeling}
\label{sec:manualannotation}

For each frame selected as described above, we cluster all points based on their 3D features and only ask the annotator to label each cluster center.

For efficiency reasons, because of the high dimension of the feature vectors, we use $k$-means clustering. The minimum value for~$k$ is the total number of considered classes, covering the ideal case where there is one cluster per class. In practice, there is a high class imbalance in point labels for driving scenes \cite{nuscenes,semantickitti,liu2022less}, which can exceed three orders of magnitude (e.g., bicycle to vegetation in SemanticKITTI). Besides, we cannot expect feature vectors from pretrained self-supervised models to provide compact, well separated clusters for each object or stuff class. As $k$-means tends to create ball-like clusters (Voronoi cells) that cover similar hyper-volumes of the feature space, we prefer to over-segment the features, letting several clusters being labeled identically. While it leads to more manual annotations, it allows us to address label-consistent feature regions that have more complex shapes in feature space. In practice, we set the number of clusters $k$ as a given, fixed fraction $\alpha$ of the number of points in the scan, making sure however that $k$ is at least an order of magnitude larger than the number of classes. This simple formulation also provide some control over the number of points to annotate, e.g., to meet an annotation budget: if the annotation budget is $N$ points and there is a total of $M$ points in all selected frames, then $\alpha \,{=}\, N/M$. Typically, we set $\alpha \,{=}\, 1\%$.

Finally, each point in a cluster receives the same label as the label to the cluster center.  
As the centroid in $k$-means clustering does not necessarily correspond to a point in the cluster, we actually take as \emph{cluster center} the point within the cluster that is the closest to the centroid, i.e., to the average point location. 
The resulting pseudo-labeled scans are then used to train a model, as described below.

\subsection{Semi-supervision}
\label{sec:teacherstudent}

We train a network in two steps. First, we exploit only the pseudo-labeled scans (with cluster-based label propagation) to train a first model. The model is obtained by minimizing the sum of the cross entropy loss $\mathcal{L}_{\rm CE}$ and the Lov\'asz loss \cite{lovasz} $\mathcal{L}_{\rm lovasz}$ between the network-predicted pointwise class probabilities $p_i$ and the pseudo-labels $y_i$, for each point $i=1, \ldots, M$ in a batch:
\begin{align}
    \label{eq:sup}
    \mathcal{L} = \frac{1}{M} \sum_{i=1}^M \lambda_{\rm CE} \; \mathcal{L}_{\rm CE}(p_i, y_i) + \lambda_{\rm lovasz} \; \mathcal{L}_{\rm lovasz}(p_i, y_i),
\end{align}
where $\lambda_{\rm CE}, \lambda_{\rm lovasz} > 0$.

In a second step, we use a teacher-student architecture with exponential moving average (EMA) to exploit the unlabeled scans as well. This second of training follows principles also exploited in, e.g., \cite{unal2023dial, distillation}. The teacher and student network parameters $\theta^t$ and $\theta^s$ are initialized with the weights obtained at the end of the first training step (see above paragraph). Then, the student parameters are further optimized by minimizing:

\begin{equation}
\begin{aligned}
\label{eq:semi}
    \mathcal{L} = \frac{1}{M} \sum_{i=1}^M \lambda_{\rm CE} \; \mathcal{L}_{\rm CE}(p^s_i, y_i) {} & + \lambda_{\rm lovasz} \; \mathcal{L}_{\rm lovasz}(p^s_i, y_i) \\ & + \lambda_{\rm KL} \; \mathcal{L}_{\rm KL}(\hat{p}^s_i, \hat{p}^t_i),
\end{aligned}
\end{equation}
%
where $\mathcal{L}_{\rm KL}$ is the KL divergence and $\lambda_{\rm KL} > 0$. Note that for scans with no pseudo-label, only the KL-divergence applies. The teacher parameters are updated after each gradient step using a exponential moving average (EMA) of the student parameters:
%
$\theta^t \leftarrow \beta \theta^t + (1 - \beta) \theta^s$.
%
In \cref{eq:semi}, $p^s_i$ is computed using a softmax layer with a temperature set to $1$, while $\hat{p}^s_i$ and $\hat{p}^t_i$ are computed using a softmax layer with a temperature $T>1$. The student is our final model. 

\subsection{Comparison to SOTA methods}
\label{sec:comparsota}

LESS~\cite{liu2022less}, which is the main competitor to our method, uses a very different approach and, more importantly, much stronger annotation assumptions. DiAL~\cite{unal2023dial} also is a strong competitor, although it typically operates 
with a two-order-of-magnitude reduction of the manually-labeled points, rather than three orders of magnitude. Like LESS, and as explained below, the provided metric in DiAL (number of manually labeled points) actually hides a higher annotation effort than the effort ordinarily implied in the domain.

LESS starts by merging consecutive scans up to a fixed number, which we don't need to do and introduces hyperparameters depending on the dataset and targeted number of segments. It then clusters points by removing the ground using a RANSAC-based approach on the cells of a pillar grid, and by grouping the remaining points using a distance threshold. Besides, large clusters are further split to favor over-segmentation over under-segmentation, and clusters that are too small are discarded. This introduces a significant number of other heuristics (as can be read in the appendix of~\cite{liu2022less}) and sensitive geometric hyperparameters. In contrast, we just create clusters using $k$-means for each frame based on point features produced by an off-the-shelf pre-trained model. To also favor over-segmentation, we just pick a large-enough~$k$ (compared to the number of classes), which is little sensitive.

In a second stage, LESS asks the human annotator to label one arbitrary point for each class in each cluster, expecting most clusters to be pure, i.e., to only contain points of a single class. In contrast, we ask the annotator only to label our cluster centers. This represents a much lower cognitive load. Indeed, when the annotator only clicks once in LESS cluster, s/he is not only labeling one point; sh/he is also guarantying that all other points in the cluster have the same label, which requires going over all points in the cluster. This extends to the case where the cluster is not pure and contains a few different classes. In contrast, when the annotator labels a point that we propose, we only require her/him to commit on a single point.

Besides, picking a point and labeling it, as in LESS, requires more operations than just labeling a proposed point, as we do. This is especially true in the case of impure segments where, in LESS, several points have to be picked and given different classes, whereas we accept the fact that a cluster can be impure and only ask for a single label on a given point, the annotator not even being aware of the existence of clusters. Consequently, the metric used in LESS (counting the number of points being manually annotated) actually hides a huge discrepancy in annotation time as well as cognitive burden, which in turn could lead to human mistakes. 

Moreover, the annotator is presented clusters constructed over accumulated scans (up to 100), which may sometimes be hard to interpret and label regarding moving objects. The reason is that these objects (e.g., cars, pedestrians) appear as long sparse tracks that are possibly colliding, i.e., overlapping.  
DiAL~\cite{unal2023dial} also leverages frame sequences by annotating discs based on aggregated scans. As performance measures related to annotation experiments in \cite{liu2022less} are purely virtual --- as are ours and most of the literature on this topic, for that matter ---, those impediments compared to a classic labeling are not observed nor measured in the annotation metric. The situation is mostly similar for DiAL, although the paper~\cite{unal2023dial}, which does not specify the number of accumulated scans, says the overhead for annotating discs of aggregated scans is 10 to 80\%. As no code is available for LESS nor DiAL, we could not evaluate this aspect. 

As a result, MILAN is significantly simpler to setup than LESS, which has 10 parameters to fix before annotating (4 of which depending on the dataset), while MILAN only has 2 ($\tau$ and~$k$).

Last, LESS uses a form of weak supervision for learning from human-informed impure clusters and an auxiliary loss term based on contrastive prototypes. It also includes the distillation of a multi-scan teacher into a mono-scan student, to produce a model processing a single frame. All this again introduces a significant number of other hyperparameters. Our method does not require all this arsenal; it is much simpler and only depends on a few little-sensitive parameters.

\section{Experiments}

This section is organized as follows. After presenting the main ingredients of our experiments (datasets, metric, backbones, \cref{sec:techdetails}), we justify our frame selection strategy. Then, we show the efficiency of our annotation and label propagation strategy. Finally, we compare \ours to state-of-the-art methods.

\subsection{Technical details}
\label{sec:techdetails}
\paragraph{Datasets and evaluation metric.}
We evaluate our method on the semantic segmentation task following \cite{wu2021redal}. 
We compare \ours to baselines on two typical 3D large-scale autonomous driving datasets: 
SemanticKITTI~\cite{semantickitti} and nuScenes~\cite{nuscenes}. 
SemanticKITTI is composed of 22 driving sequences. 
Following the official protocol, we evaluate on the validation split (seq 08) and train the models on the entire official training split (seq 00-07 and 09-10).
We also evaluate on nuScenes, which is composed of scenes acquired in Boston and Singapore and annotated with 16 semantic classes. We use the official split with 700 scenes for training and 150 scenes for validation. 
In all our experiments, we employ the mIoU to evaluate the semantic segmentation quality.

\paragraph{Frame selection.} 
Following SeedAL~\cite{samet2023seedal}, we represent each scan by averaging its point features obtained from self-supervised 3D features (i.e. ScaLR). Next, we prune the SemanticKITTI and nuScenes datasets using $\tau$ thresholds of 0.95 and 0.92, respectively. By applying these thresholds, we were able to reduce the size of the SemanticKITTI dataset to approximately 12\% of its original size, while the nuScenes dataset was reduced to 50\% of its initial size. We use the features associated to each scan to calculate the cosine similarity between two scans.

\paragraph{Network architectures.} Following previous works~\cite{wu2021redal, samet2023seedal}, we train the 3D segmentation model SPVCNN~\cite{spvnas} which is based on point-voxel CNN. We also employ the recent WaffleIron~\cite{puy23waffleiron} with feature size 768 (WI-768), which achieves good results with little annotation in particular when pre-trained in a self-supervised fashion \cite{puy2024scalr}.

\paragraph{Network training.} During the first training step using only pseudo-labeled scans, we use $\lambda_{\rm CE} = \lambda_{\rm lovasz} = 1$ when training a WaffleIron backbone, and $\lambda_{\rm CE} = 0$, $\lambda_{\rm lovasz} = 1$ when training a SPVCNN backbone. During the second training step using the teacher-student architecture, we use $\beta=0.99$, $\lambda_{\rm CE} = 0.5$, $\lambda_{\rm lovasz} = 1$, $\lambda_{\rm KL} = 0.5 T^2$ and a temperature $T = 4$ in the KL-divergence loss. Note that in our implementation, the teacher and the student receives the same point cloud. We activated the stochastic depth~\cite{stocdepth} layers in WaffleIron to enable variations between the student and teacher outputs. Unless otherwise mentioned, any training involving WaffleIron and less than $100\%$ of manually annotated data leverages the pretrained weights obtained with ScaLR \cite{puy2024scalr}. The score obtained with $100\%$ of manually annotated data are obtained without pretraining.

\begin{table}[h]
    \caption{Performance of semantic segmentation (mIoU\%) when propagating labels after clustering
    using only 1\% of nuScenes's validation set. 
    (Please note that this table reports mIoU while \cref{table:purity_nuscenes} reports mean accuracy.)}
    \begin{center}
    \setlength{\tabcolsep}{3pt}
    \begin{tabular}{lccc}
    \toprule
       Method & ScaLR & Openscene & BEVContrast \\
        \midrule
        mIoU\% & \textbf{75.5} & 68.9 & 56.9  \\
    \bottomrule  
    \end{tabular}
    \end{center}
    \label{tab:ss3dfeats}
\end{table}

\subsection{Selection of efficient self-supervised 3D features}
\label{sec:select3Dfeats}

3D point features are used at multiple stages of our pipeline: frame selection, label propagation and pretraining.
To determine the best self-supervised 3D features for our use, we conducted experiments on label propagation, which is the most sensitive stage of our pipeline. We used three different self-supervised 3D features: (i)~ScaLR \cite{puy2024scalr} features distilled from DINOv2 \cite{oquab2023dinov2} via images, (ii)~OpenScene \cite{peng2023openscene} features distilled from CLIP \cite{radford2021clip} via images, and (iii)~pure 3D features of BEVContrast \cite{sautier2024bevcontrast} originating from unsupervised contrastive learning on point clouds. As shown in Table \ref{tab:ss3dfeats}, ScaLR features significantly outperformed the others. Therefore, we chose ScaLR features for all our pipeline.  
Please refer to the Appendix for more results.

\begin{table*}[!t]
\caption{\textbf{Classwise accuracy (\%) of our pseudo-labels on SemanticKITTI}. For each class, we compute the ratio between the number of correct pseudo-labels divided by the ground truth number of points in this class.}
\label{table:purity}
\scriptsize
\ra{1.5}
\newcommand*\rotext{\multicolumn{1}{R{90}{0mm}}}
\setlength{\tabcolsep}{1.2pt}
\centering
\begin{tabular}{l| l | r r r r r r r r r r r r r r r r r r }
\toprule 
\rotext{\% clicks}
    & \rotext{Average}
    & \rotext{car}
    & \rotext{bicycle}
    & \rotext{motorcycle}
    & \rotext{truck}
    & \rotext{other-vehicle}
    & \rotext{person}
    & \rotext{bicyclist}
    & \rotext{road}
    & \rotext{parking}
    & \rotext{sidewalk}
    & \rotext{other-ground}
    & \rotext{building}
    & \rotext{fence}
    & \rotext{vegetation}
    & \rotext{trunk}
    & \rotext{terrain}
    & \rotext{pole}
    & \rotext{traffic-sign}
\\
\midrule
0.01
    & 87.2
    & 97.4%
    & 38.0%
    & 91.8%
    & 98.9%
    & 97.4%
    & 85.4%
    & 85.5%
    & 96.8%
    & 87.6%
    & 91.1%
    & 90.6%
    & 98.1%
    & 88.8%
    & 95.7%
    & 83.8%
    & 95.7%
    & 62.1%
    & 85.1%
\\
0.05
    & 87.3
    & 96.9%
    & 48.0%
    & 93.9%
    & 98.1%
    & 97.0%
    & 84.1%
    & 87.6%
    & 96.6%
    & 87.1%
    & 92.1%
    & 90.4%
    & 97.5%
    & 89.1%
    & 96.1%
    & 81.3%
    & 93.8%
    & 62.6%
    & 80.1%
\\
0.1 
    & 87.2
    & 96.6%
    & 51.3%
    & 93.0%
    & 98.0%
    & 97.0%
    & 84.4%
    & 85.8%
    & 96.5%
    & 87.4%
    & 92.3%
    & 89.2%
    & 97.2%
    & 89.3%
    & 96.0%
    & 82.1%
    & 91.7%
    & 62.3%
    & 79.5%
\\
\bottomrule
\end{tabular}
\end{table*}

\begin{table*}[!t]
\caption{\textbf{Classwise accuracy (\%) of our pseudo-labels on nuScenes}. For each class, we compute the ratio between the number of correct pseudo-labels divided by the ground truth number of points in this class.}
\label{table:purity_nuscenes}
\scriptsize
\ra{1.5}
\newcommand*\rotext{\multicolumn{1}{R{90}{0mm}}}
\setlength{\tabcolsep}{2pt}
\centering
\begin{tabular}{l |l | r r r r r r r r r r r r r r r r}
\toprule 
\rotext{\% clicks}
    & \rotext{Average}
    & \rotext{barrier}
    & \rotext{bicycle}
    & \rotext{bus}
    & \rotext{car}
    & \rotext{const. veh.}
    & \rotext{motorcycle}
    & \rotext{pedestrian}
    & \rotext{traffic cone}
    & \rotext{trailer}
    & \rotext{truck}
    & \rotext{driv. surf.}
    & \rotext{other flat}
    & \rotext{sidewalk}
    & \rotext{terrain}
    & \rotext{manmade}
    & \rotext{vegetation}
\\
\midrule
0.2
    & 89.5
    & 93.1%
    & 65.6%
    & 99.1%
    & 96.0%
    & 92.9%
    & 90.2%
    & 72.3%
    & 63.9%
    & 98.0%
    & 97.9%
    & 97.8%
    & 90.3%
    & 91.1%
    & 93.3%
    & 96.3%
    & 94.1%
\\
0.6
    & 89.1%
    & 93.9%
    & 61.1%
    & 98.6%
    & 95.1%
    & 92.7%
    & 89.9%
    & 73.2%
    & 65.6%
    & 97.1%
    & 96.8%
    & 97.9%
    & 90.5%
    & 90.9%
    & 92.3%
    & 95.3%
    & 94.3%
\\
0.9 
    & 88.6%
    & 93.7%
    & 59.5%
    & 98.2%
    & 94.6%
    & 92.1%
    & 88.2%
    & 73.4%
    & 65.4%
    & 96.5%
    & 96.2%
    & 97.9%
    & 90.3%
    & 90.9%
    & 91.7%
    & 94.8%
    & 94.4%
\\
\bottomrule
\end{tabular}
\end{table*}
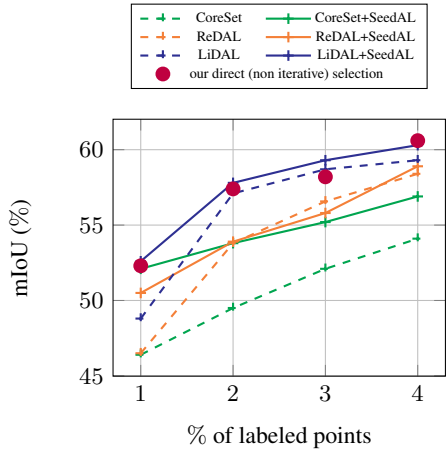
\begin{figure}[h]
\centering
\begin{tikzpicture}
\tikzstyle{every node}=[font=\small]
\begin{axis}[
    width=6cm,
    height=5cm,
    font=\footnotesize,
    xtick={1,2,3,4},
    ylabel=mIoU ($\%$),
    xlabel=$\%$ of labeled points,
    ylabel shift=-15 pt,
    xlabel shift=-15 pt,
    tick label style={font=\small},
    legend columns=2, 
    grid=major,
    legend style={
        nodes={scale=0.6, transform shape}, 
        at={(0.5,1.45)},
        anchor=north,
        /tikz/column 2/.style={
            column sep=5pt,
        },
    },
]
\addplot[thick, dashed, color=Green, mark=+] coordinates {(1, 46.4) (2, 49.5) (3, 52.1) (4, 54.1)};
\addlegendentry{CoreSet};
\addplot[thick, color=Green, mark=+] coordinates {(1, 52.1) (2, 53.8) (3, 55.2) (4, 56.9)};
\addlegendentry{CoreSet+SeedAL};
\addplot[thick, dashed, color=Orange, mark=+] coordinates {(1, 46.5) (2, 53.8) (3, 56.57) (4, 58.4)};
\addlegendentry{ReDAL};
\addplot[thick, color=Orange, mark=+] coordinates {(1, 50.5) (2, 53.9) (3, 55.8) (4, 58.9)};
\addlegendentry{ReDAL+SeedAL};
\addplot[thick, dashed, color=Blue, mark=+] coordinates {(1, 48.8) (2, 57.1) (3, 58.7) (4, 59.3)};
\addlegendentry{LiDAL};
\addplot[thick, color=Blue, mark=+] coordinates {(1, 52.6) (2, 57.8) (3, 59.3) (4, 60.3)};
\addlegendentry{LiDAL+SeedAL};
\addplot[color=purple, ultra thick, mark=* , mark size=2pt, only marks] coordinates {(1, 52.3) (2, 57.4) (3, 58.2) (4, 60.6)};
\addlegendentry{our direct \rlap{(non iterative) selection}};
\end{axis}
\end{tikzpicture}
\caption{Comparison of our direct (non iterative) data selection strategy \ours with SOTA active learning strategies on SemanticKITTI. All models are trained using SPVCNN as network architecture and using solely the scans with labeled points for training (no semi-supervision). 
}
\label{fig:scan_selection}
\end{figure}

\subsection{Direct scan selection using self-supervised 3D features}
\label{sec:scanselect}

The first step of our method consists of a selection of scans to be annotated. For our method to be annotation efficient, training a model on these scans must lead to high performance. 
We show our scan selection method is highly competitive by comparing it to state-of-the-art active learning methods: CoreSet~\cite{coreset}, ReDAL~\cite{wu2021redal}, LiDAL~\cite{hu2022lidal}, and SeedAL~\cite{samet2023seedal}. Note that competing with such active learning method is challenging. Indeed, if the budget of scans to select corresponds, e.g., to 4\% of the considered training set, active learning methods will first select a smaller amount of points, e.g., 1\% of the training set, fully annotate these points, train a model with these annotated points, and use this model to select a new subset of points. This process is repeated until the annotation budget is exhausted. Instead, we keep the annotation cost to \emph{zero} for our selection. We do so by leveraging self-supervised 3D features (see \cref{sec:frameselection}).

We present in \cref{fig:scan_selection} the performance reached by training SPVCNN using our selection of 1\%, 2\%, 3\% and 4\% of the training set of SemanticKITTI. We notice that our selection of scans allows us to reach a performance within $100.5\%$ and $96.5\%$ of the performance obtained by the best active learning method, i.e., LiDAL initialized with SeedAL~\cite{hu2022lidal,samet2023seedal}. We recall that, for this experiment, all methods use full manually-annotated scans. We show in the next section that these selected scans can be cheaply annotated.

\subsection{Cheap annotation of selected scans}

The principle of our efficient annotation technique is simple: cluster points which are semantically similar, select one point in each cluster, ask the annotator to label the points, propagate the labels in each cluster. To reach an annotation budget of $x$ percent of a given training set ($x\%$ of clicks), we use the following protocol: (a)~we select a subset of $100\,x$ percent of scans with our scan selection method; (b)~for each of these selected scans, we cluster the points  with a number of clusters equal to $1\%$ of the number of points. 
The same protocol applies to both SemanticKITTI and nuScenes, but for nuScenes, which has more diversity than SemanticKITTI, we double the number of clusters per selected scan, leading to $2 x\%$ of clicks.

We start by showing that the clusters we obtain by clustering ScaLR \cite{puy2024scalr} features permits us to obtain accurate pseudo-labels. We report in \cref{table:purity} the classwise accuracy computed over our selected set of scans: number of correct pseudo-labels in a given class divided by ground truth number of points in this class. We notice that our label propagation strategy works well as, on average over all classes, more that 87\% of the points in each class receive good pseudo-labels. We remark nevertheless that the classes that are rare and correspond to small objects, like bicycle and pole, are more difficult to pseudo-label accurately. Note also that we removed the class `motorcyclist' (a very rare class in SemanticKITTI) in \cref{table:purity} as it never appeared in our selected scans. We report the same numbers on nuScenes in \cref{table:purity_nuscenes} and draw similar conclusions. 

\begin{table*}[t]
\ra{1.1}
\setlength{\tabcolsep}{5pt}
\caption{
Performance reached using $5.0\%$ of scans selected using our strategy and using only $0.05\%$ of annotations, i.e., $1\%$ of annotated points in the selected scans, which we propagate within each cluster.
}
\label{table:training_step}
\centering
\begin{tabular}{lcccccccc}
\toprule
Backbone 
    & ScaLR pretrained
    & Teacher-Student
    & \%\,labels 
    & \%\,scans 
     
    & mIoU\% 
    
\\
\midrule

WI-768 
    & \xmark
    & \xmark
    & 100 
    & 100 
    & 63.4 
\\ 
\midrule
WI-768 
    & \xmark 
    & \xmark 
    & 100 
    & 5 
    & 61.0 
\\  
WI-768 
    & \cmark 
    & \xmark 
    & 100 
    & 5 
    & 63.0 
\\  
WI-768 
    & \cmark 
    & \cmark 
    & 100 
    & 5 
    & 63.3 
\\ 
\midrule
WI-768 
    & \xmark 
    & \xmark 
    & 1 
    & 5 
    & 57.2 
\\  
WI-768 
    & \cmark 
    & \xmark
    & 1
    & 5 
    & 61.2 
\\ 
WI-768 
    & \cmark 
    & \cmark 
    & 1
    & 5 
    & 62.9 
\\ 
\bottomrule
\end{tabular}
\end{table*}

\subsection{Ablation study}

The main ingredients of our method are: relevant frame selection, feature-based point clustering, pretraining and semi-supervised training with a teacher-student approach. The first three stages rely on pre-trained 3D point features.

\paragraph{3D point features.}

The choice of ScaLR \cite{puy2024scalr} to produce 3D point features is discussed in \cref{sec:select3Dfeats}.

\paragraph{Frame selection.}

The relevance of our frame selection is covered in \cref{sec:scanselect}, where we show we are better or on par with state-of-the-art active learning methods in their best configuration, i.e., when they rely on a state-of-the-art initialization.

\paragraph{Label propagation.}

As for the clustering-based annotation, we tried replacing features by point coordinates, loosing -8.5 mIoU pts at 0.1\% of labels with nuScenes. This shows that basic geometric features are much less powerful than the more semantic features obtained from WaffleIron (WI-768) pretrained with ScaLR.

\paragraph{Pretrained features and finetuning.}

On SemanticKITTI, using pretrained ScaLR features and finetuning with the propagated labels gains +2.0 mIoU pts at 5\,\% labels and +4.0 pts at 0.05\%, as shown in \cref{table:training_step}.
We also notice that this finetuning of WI-768 on our selection of 5\% of scans reaches a similar performance as a non-pretrained WI-768 trained on 100\% of data. 

As another token of comparison, just finetuning on 1\% of labels in nuScenes gets 50.7\% mIoU \cite{puy2024scalr}, while with only 0.9\% of labels, which we then propagate, we reach 76.2\% mIoU after just finetuning.

\paragraph{Semi-supervision via teacher-student training.}

We include a teacher-student training to exploit unlabeled data and improve segmentation quality. Benefits of applying a teacher-student shows in \cref{table:training_step}.
On SemanticKITTI, after teacher-student training from 1\% of labels in 5\% of scans, we reach nearly the same performance as a non-pretrained WI-768 trained on 100\% of data. We also gain +1.7 mIoU pts at 0.1\% labels on SemanticKITTI (cf.\ \cref{tab:resultsSK}) and +1.0 pts at 0.9\% on nuScenes (cf.\ \cref{tab:resultsNS}).
However, as our teacher-student training currently is very basic, we sometimes degrade the performance at very low labeling ratios: -0.3\,mIoU pts at 0.01\% labels on SemanticKITTI and -0.3\,pts at 0.2\% labels on nuScenes. A more robust teacher-student mechanism is future work.

\newcommand\tabsetting{
\tabcolsep 2pt
\newcommand\D{\hphantom{0}}
\newcommand\DD{\D\D}
\newcommand\n[1]{\textsuperscript{(##1)}}
\newcommand\seedal{\cite{samet2023seedal}}
\newcommand\spvnas{\cite{spvnas}}
\newcommand\spvcnn{\cite{spvnas}}
\newcommand\dial{\cite{unal2023dial}}
\newcommand\laser{\cite{kong2023lasermix}}
\newcommand\coars{\cite{li2022coarse3d}}
\newcommand\sscn{\cite{graham2018spconv}}
\newcommand\minknet{\cite{choy2019minknet}}
\newcommand\annot{\cite{xie2023annotator}}
\newcommand\consc{\cite{hou2021contrastsc}}
\newcommand\cyltd{\cite{cylinder3d}}
\newcommand\less{\cite{liu2022less}}
\newcommand\lidal{\cite{hu2022lidal}}
\newcommand\otoc{\cite{liu2021otoc}}
\newcommand\redal{\cite{wu2021redal}}
\newcommand\sqn{\cite{hu2022sqn}}
\newcommand\randla{\cite{hu2021learning}}
\newcommand\waffle{\cite{puy23waffleiron}}
\newcommand\scalr{\cite{puy2024scalr}}
\newcommand\salsa{\cite{cortinhal2020salsanext}}
\newcommand\basal{\cite{wei2023basal}}
\newcommand\hybrid{\cite{li2022hybridcr}}
\newcommand\tmsgp{\cite{shi2022weakly}}
\newcommand\spconv{\cite{graham2018spconv}}
\newcommand\wstd{\cite{liu2022ws3d}}
\newcommand\wstdpp{\cite{liu2023generalized}}
\newcommand\clipts{\cite{chen2023clip2scene}}
\newcommand\seal{\cite{liu2023seal}}
\newcommand\na{\mbox{\scriptsize\color{gray} N/A}}
\newcommand\rf[2]{\,{\scriptsize ##1\,'##2}}
\newcommand\IN{\,in\!}
\newcommand\dg{{$^\dagger$}}
}

\begin{table*}[t]
\centering
\tabsetting
\caption{Performance on SemanticKITTI val set (\%mIoU), for various proportions $x$ of manually-labeled points.
*Incomparable annotation metric (underestimated).
\dg Overestimated.
\n{1}Operating on single scan. 
\n{a}Operating on aggregated scans.
\n{n}No teacher-student used.}
\label{tab:resultsSK}
\begin{tabular}{l@{}l@{}lcclll}
& & \!\!~$x\%$\!\! & \%mIoU & \%mIoU & mIoU &   \\
Method & Reference & \!\!labels\!\! & @\,$x$\% & @\,100\%  & ratio & Backbone \\
\midrule
LaserMix\n{vox}& \laser\rf{cvpr}{23} & \D 5.0       & 56.7 & \na  & \D \na  & Cylinder3D \cyltd\IN\laser \\
DiAL\n{1}  & \dial\rf{ra-l}{23}  & \D 5.0       & 58.1 & 63.8 & \D 91.1 & SPVCNN \spvcnn \\
ReDAL    & \redal\rf{iccv}{21} & \D 5.0       & 59.8 & 61.4 & \D 97.4 & MinkUNet \minknet \\
LiDAL    & \lidal\rf{eccv}{22} & \D 5.0       & 60.1 & 61.4 & \D 97.9 & MinkUNet \minknet \\
\rowcolor{green!20}
\ours    & (ours)          & \D 5.0       & \textbf{63.3} & 63.4 & \D \textbf{99.8} & WaffleIron \waffle\IN\scalr \\
\rowcolor{gray!40}
*DiAL\n{a}  & \dial\rf{ra-l}{23}          & \D \llap*5.0       & \llap\dg63.8 & 63.8 & \llap\dg100.0 & SPVCNN \spvcnn \\
\midrule
WS3D   & \wstd\rf{eccv}{22}    & \D 1.0       & 38.9 & 66.9 & \D 58.1 & SPConv \spconv \\
CLIP2Scene & \clipts\rf{cvpr}{23} & \D 1.0   & 42.6 & 55.0 & \D 77.5 & MinkUNet \minknet \\
WS3D++ & \wstdpp\rf{arxiv}{23} & \D 1.0       & 46.8 & 76.8 & \D 60.9 & SPConv \spconv \\
HybridCR & \hybrid\rf{cvpr}{22}& \D 1.0       & 51.9 & 53.2 & \D \textbf{97.6} & HybridCR \hybrid \\
\rowcolor{gray!40}
*DiAL\n{a} & \dial\rf{ra-l}{23} & \D \llap*1.0       & \llap\dg61.4 & 63.8 & \D \llap\dg96.2 & SPVCNN \spvcnn \\
\midrule
ContrastSC&\consc\IN\less\rf{cvpr}{21}& \D 0.1       & 46.0 & \na  & \D \na  & MinkUNet \minknet \\
Annotator& \annot\rf{neurips}{23}     & \D 0.1       & 53.7 & \na  & \D \na  & MinkUNet \minknet \\
SQN      & \sqn\rf{eccv}{22}          & \D 0.1       & 55.8 & \na  & \D \na  & MinkUNet \minknet\IN\sqn \\
COARSE3D & \coars\rf{bmvc}{22}        & \D 0.1       & 56.6 & 58.4 & \D {96.9} & SalsaNext \salsa\IN\coars \\
\rowcolor{green!20}
\ours    & (ours) & \D 0.1       & \textbf{63.1} & 63.4 & \D \textbf{99.5} & WaffleIron \waffle\IN\scalr \\
\rowcolor{gray!40}
*LESS\n{a} & \less\rf{eccv}{22} & \D \llap*0.1  & \llap\dg66.0 & 65.9 & \llap\dg100.2 & Cylinder3D \cyltd\IN\less \\
\midrule
COARSE3D & \coars\rf{bmvc}{22}       & \D 0.01      & 47.1 & 58.4 & \D 80.7 & SalsaNext \salsa\IN\coars \\
SQN      & \sqn\rf{eccv}{22}         & \D 0.01      & 50.0 & \na  & \D \na  & MinkUNet \minknet \\
\rowcolor{green!20}
\ours    & (ours) & \D 0.01      & \textbf{\textit{58.2\n{n}}} & 63.4 & \D \textit{\textbf{91.8}} & WaffleIron \waffle\IN\scalr \\
\rowcolor{gray!40}
*LESS\n{a} & \less\rf{eccv}{22}  & \D \llap*0.01 & \llap\dg61.0 & 65.9 & \D \llap\dg92.6 & Cylinder3D \cyltd\IN\less \\
\end{tabular}
\end{table*}

\subsection{Annotation effort}
\label{sec:annoteffort}

What ultimately matters is the cost of manually annotating a given fraction of the dataset, leading to a certain level of performance. As experimenting with human annotators is costly and difficult to set up, the proportion of labeled points is generally assumed to be representative of the annotation cost. However, annotation procedures differ from one method to another, and the same percentage of manually-labeled points may actually correspond to widely different annotation costs. In this section, we study the annotation effort with MILAN, compared to the annotation effort with LESS, which is our main competitor.

\paragraph{Differing annotation procedures.}

LESS \cite{liu2022less} partitions aggregated scans into components. The annotators' task is to thoroughly examine each component, searching for \emph{all classes} that are present in each, and randomly label $n$ points per class whose proportion of points in the component is more than $p$, where parameter $n$ depends on the overall labeling ratio and parameter $p$ depends on the dataset \cite[Sec.\,S.1]{liu2022less}.
LESS annotation time thus depends on component size and, more importantly, on their purity, while 1/5 to 1/3 of segments are impure \cite[Tab.\,5]{liu2022less}.

In contrast, we do not show segments; we show the whole point cloud and ask annotators to choose \emph{in context} a label for a few \emph{highlighted, selected points}. Our task is much lighter as we need no exhaustive search of all classes in each component, nor any minimum proportion assessment. 
As our selected points are cluster centers (with purity $\sim$88\%, \cref{table:purity,table:purity_nuscenes}), labeling decision is even easier, compared to checking the class of points located at component boundaries, as in LESS.

Besides, with LESS \cite{liu2022less} as well as DiAL \cite{unal2023dial}, the annotator is presented areas of point clouds aggregated over several scans. In this way, more points can be labeled at once and annotation can be easier thanks to point densification, especially in regions that would be sparse in individual scans (e.g., far from the sensor). However, annotation can also become harder regarding moving objects (cars, pedestrians, etc.) that are then represented as diluted traces in space, that additionally may overlap.
Although we did not experiment it, scan accumulation is also possible with MILAN, to provide the annotator with more context and more details while keeping the same selected points to label.

We tried leveraging aggregation to further reduce the labeling ratio, but so far got little improvement. It is probably because successive scans are highly redundant, which our pruning already deals with. Also, we rely on the rich features of ScaLR, which allow us to generalize well from a small number of annotated scans.

\paragraph{Quantitative comparison.}

Comparing actual annotation times is hard as LESS only does virtual experiments and provides no code.
Still, we had a knowledgeable annotator label one scene from SK in a LESS-like fashion, i.e., after separating and chunking the ground, and creating TARL segments \cite{tarl}, which are very much alike LESS segments. The annotation time for LESS, without even asking for class proportion checking, was 5.3\,h, vs 1.8\,h for MILAN (about $3\times$ less). The reason is that, when annotating in LESS way, although there are less clusters to label, a lot of time is spent searching for all possible classes in each segment.

Arguably, annotating 0.1\% of points in the LESS way has the same human cost as annotating 0.3\% of points ($3\times$ more points) in the MILAN way. In other words, when LESS segmentation results are provided for a given proportion of 0.1\% of labeled points, it actually has to be understood as about 0.3\% of labeled points when comparing to results with points annotated in the MILAN way.

Nevertheless, in what follows, results are reported relatively to the actual proportion of manually-labeled points, not relatively to an (approximate) human annotation cost. But one has to keep in mind that the proportion of labeled points in LESS is, in way, underestimated by about a factor of 3 as a measure of the annotation effort; likewise, the corresponding performance of LESS is overestimated.

\begin{table*}[t]
\centering
\tabsetting
\caption{Performance on nuScenes val set (\%mIoU) for various proportions $x$ of manually-labeled points. 
*Incomparable annotation metric (underestimated).
\dg Overestimated.
\n{a}Operating on aggregated scans.
\n{n}No teacher-student used.
}
\label{tab:resultsNS}
\begin{tabular}{l@{}llcclll}
& & \!\!~$x\%$\!\! & \%mIoU & \%mIoU & mIoU &   \\
Method & Reference & \!\!labels\!\! & @\,{$x$}\% & @\,100\%  & ratio & Backbone \\
\midrule
ReDAL & \redal\rf{iccv}{21} & \D 5.0       & 58.3 & 71.7 & \D 81.3 & SPVCNN \spvcnn \\
LiDAL & \lidal\rf{eccv}{22} & \D 5.0       & \textbf{68.2} & 71.7 & \D \textbf{91.2} & SPVCNN \spvcnn \\
\midrule
WS3D   & \wstd\rf{eccv}{22}    & \D 1.0       & 49.1 & 71.6 & \D 68.6 & SPConv \spconv \\
ScaLR   & \scalr\rf{cvpr}{24}    & \D 1.0       & 51.0 & 78.4 & \D 65.1 & WaffleIron \waffle \\
WS3D++ & \wstdpp\rf{arxiv}{23} & \D 1.0       & 53.7 & 77.8 & \D 69.0 & SPConv \spconv \\
LaserMix\n{vox}& \laser\rf{cvpr}{23} & \D 1.0     & 55.3 & 74.1 & \D 74.6 & Cylinder3D \cyltd\IN\laser \\
CLIP2Scene & \clipts\rf{cvpr}{23}  & \D 1.0   & \textbf{56.3} & 71.5 & \D \textbf{78.7} & MinkUNet \minknet \\
Seal   & \seal\rf{neurips}{23} & \D 1.0   & 45.8 & 75.6 & \D 60.6 & MinkUNet \minknet \\
\midrule
ContrastSC & \consc\IN\less\rf{cvpr}{21} & \D 0.9       & 65.5 & \na & \na  & MinkUNet \minknet \\
\rowcolor{green!20}
\ours & (ours)              & \D 0.9      & \textbf{77.2}~ & 78.7 & \D \textbf{98.0} & WaffleIron \waffle\IN\scalr \\
\rowcolor{gray!40}
*LESS\n{a} & \less\rf{eccv}{22} & \D \llap*0.9 & \llap\dg74.8 & 75.4 & \D \llap\dg99.2 & Cylinder3D \cyltd\IN\less \\
\midrule
ContrastSC & \consc\IN\less\rf{cvpr}{21}  & \D 0.2       & 63.5 & \na & \na  & MinkUNet \minknet \\
\rowcolor{green!20}
\rowcolor{green!20}
\ours & (ours) & \D 0.2      & 72.3\relax{\n{n}} & 78.7 & \D 91.9 & WaffleIron \waffle\IN\scalr \\
\rowcolor{gray!40}
*LESS\n{a}  & \less\rf{eccv}{22}         & \D \llap*0.2 & \llap\dg73.5 & 75.4 & \D \llap\dg97.5 & Cylinder3D \cyltd\IN\less \\
\midrule
COARSE3D  & \coars\rf{bmvc}{22} & \D 0.1       & 58.7 & 72.2 & \D 81.3 & SalsaNext \salsa \\
\rowcolor{green!20}
\rowcolor{green!20}
\ours     & (ours) & \D 0.1 & {70.6}\relax{\n{n}} & 78.7 & \D 89.7   & WaffleIron \waffle\IN\scalr \\
\midrule
TMSGP & \tmsgp\rf{cvpr}{22} & \D0.00\rlap1 & 50.3 & 60.7 & \D 82.9 & MinkUNet \minknet \\
\end{tabular}
\end{table*}

\subsection{Comparison to the state of the art}

\paragraph{Baselines.} We now evaluate the performance of \ours on both SemanticKITTI \cite{semantickitti} and nuScenes \cite{nuscenes}, %
and compare it against the performance of state-of-the-art methods which aim at reducing annotation costs using different paradigms. We consider: 
few-shot methods (Seal \cite{liu2023seal}, CLIP2Scene \cite{chen2023clip2scene}, ScaLR \cite{puy2024scalr}, ContrastSC \cite{hou2021contrastsc});
recent active learning strategies (ReDAL \cite{wu2021redal}, LiDAL \cite{hu2022lidal}, SeedAL \cite{samet2023seedal}, Annotator \cite{xie2023annotator}); 
semi-supervised methods (LaserMix \cite{kong2023lasermix}, COARSE3D~\cite{li2022coarse3d}, HybridCR \cite{li2022hybridcr}, WS3D \cite{liu2022ws3d}, SQN \cite{hu2022sqn});
efficient annotation strategies like \ours (LESS \cite{liu2022less}, DiAL \cite{unal2023dial}, TMSGP \cite{shi2022weakly}).

Note that, as  LESS \cite{liu2022less} and DiAL \cite{unal2023dial}
are not directly comparable to \ours because one point labeling requires more effort for the annotator, who needs: (1)~to handle moving objects in the multi-scan representation and, for LESS, as detailed in \cref{sec:annoteffort}, (2)~to find \emph{all} classes present in a cluster, which requires to check the whole cluster and possibly label several points, i.e., one per class, and (3)~to select a point \emph{and} a class per label, rather than just provide the class of a given highlighted point. Besides, neither LESS nor DiAL have available code, which prevents providing other numbers that those provided in their respective papers \cite{liu2022less, unal2023dial}.

\paragraph{Metrics.} 

For a given proportion of manually-labeled points (ignoring whether point annotation is simpler or harder for the given methods), we report the mIoU score of the resulting semantic segmentation.

As stressed in \cref{tab:resultsSK,tab:resultsNS}, there is no consensus on a backbone for evaluation. Worse, even when backbones are seemingly equal, their performance at 100\% labels may widely differ. As some methods have no code, it is hard to evaluate backbone impact. To reach our milli-annotation goal, we had to use a SOTA 3D backbone with SOTA pretraining. Yet, to mitigate backbone variety, we measure the relative performance to full supervision too: we report the relative mIoU ratio when comparing to the mIoU resulting from a training with the fully-annotated dataset (100\%).

\paragraph{SemanticKITTI results.} 
We present the results obtained with \ours and all baselines on SemanticKITTI in \cref{tab:resultsSK}. 
First, as expected, we observe that best performing methods use semi-supervision to leverage at best the little amount of available annotations and the large amount of unannotated data. Second, we observe that with 5\% and 0.1\% of annotated data, \ours\ permits us to reach about 99.0\% of the performance we would have obtained by training WI-768 on the fully manually annotated training set.  
Note that the reference mIoU with full supervision that we provide (63.8\%) is averaged over two trainings, as we noticed variability between different trainings of WI-768 on the full training set. Let us mention as well that the score we provide for \ours at 0.01\% of labeled points is obtained using few-shot finetuning but without semi-supervision with the teacher-student technique, which did not lead to a better score. Overall, the results we reached with \ours are equivalent to those of the method LESS, which however requires much more annotation effort (see \cref{sec:comparsota}).

\paragraph{nuScenes results.} We present the results obtained with \ours and all baselines on nuScenes in \cref{tab:resultsNS}. Let us mention that the score we provide for \ours at 0.2\% of labeled points is obtained using few-shot finetuning but without semi-supervision with the teacher-student technique, which did not lead to a better score. We believe that the proportion of pseudo-labeled points with respect to unannotated points is too small to provide sufficient guidance is this low annotated data regime. We leave improvement of the teacher-student approach for future work. Nevertheless, we notice that we are able to reach an mIoU of $77.2\%$ with as little as 0.9\% percent of annotations, even surpassing by a large margin methods using $5\%$ of annotated data.

\paragraph{Discussion.}

As shown with the above experiments, MILAN are in general on par with the SOTA, sometimes even better, while using a much simpler annotation strategy, and despite a misleading metric in terms of \% of labeled points.

DiAL is somehow designed for ``centi-annotation'', i.e., reducing the annotation effort by a factor of $10^2$. While MILAN is on par with DiAL at 5\% labels (-0.5 pts, which is within training variation), MILAN outperforms DiAL at lower ratios: we gain +1.7 pts at 0.1\% labels, while using 10$\times$ less labels (Tab.\,4).

As for LESS, it definitely targets ``milli-annotation'', i.e., reducing the annotation effort by a factor of $10^3$. In our experiments, MILAN largely outperform LESS in mIoU at 0.9\% labels (Tab.\,5) and is similar to LESS ($\sim$\,1 pt below) on the ratio to mIoU at 100\% labels in most cases. Moreover, we argued above that the annotation effort in MILAN is significantly lighter for the same amount of points to manually label. It means the actual performance of LESS is overestimated, when comparing to MILAN.

MILAN is also largely simpler than DiAL and LESS, which paves the way for further improvements. LESS has 10 geometric/time hyperparameters (besides training), 4 of which significantly depend on the dataset; MILAN only has two ($\tau$, $k$), where $\tau$ only mildly depends on the dataset.

\section{Conclusion}

We have presented a simple and highly annotation-efficient method, which leverages at best available powerful self-supervised features and permits us to reach a semantic segmentation performance nearly as good as with a fully-annotated dataset, while only requiring about one thousandth of the point labels. 

\section*{Acknowledgments}
This work was granted access to HPC resources
of IDRIS under GENCI allocation 2024-AD011014946. This research also received the support of EXA4MIND project, funded by a European Union's Horizon Europe Research and Innovation Programme, under Grant Agreement N° 101092944. Views and opinions expressed are however those of the author(s) only and do not necessarily reflect those of the European Union or the European Commission. Neither the European Union nor the granting authority can be held responsible for them. We acknowledge EuroHPC Joint Undertaking for awarding us access to Karolina at IT4Innovations, Czech Republic (Projects EU2023D03-040 and EU2023R02-128).

\bibliographystyle{ieee_fullname}
\bibliography{main}

\begin{thebibliography}{10}\itemsep=-1pt

\bibitem{Ash2020Deep}
Jordan~T. Ash, Chicheng Zhang, Akshay Krishnamurthy, John Langford, and Alekh Agarwal.
\newblock Deep batch active learning by diverse, uncertain gradient lower bounds.
\newblock In {\em International Conference for Learning Representations (ICLR)}, 2020.

\bibitem{semantickitti}
Jens Behley, Martin Garbade, Andres Milioto, Jan Quenzel, Sven Behnke, Cyrill Stachniss, and Jurgen Gall.
\newblock {{SemanticKITTI}: A Dataset for Semantic Scene Understanding of LiDAR Sequences}.
\newblock In {\em International Conference on Computer Vision (ICCV)}, pages 9297--9307, 2019.

\bibitem{lovasz}
Maxim Berman, Amal~Rannen Triki, and Matthew~B Blaschko.
\newblock The lov{\'a}sz-softmax loss: A tractable surrogate for the optimization of the intersection-over-union measure in neural networks.
\newblock In {\em Conference on Computer Vision and Pattern Recognition (CVPR)}, 2018.

\bibitem{boulch2023also}
Alexandre Boulch, Corentin Sautier, Björn Michele, Gilles Puy, and Renaud Marlet.
\newblock {ALSO}: {Automotive} lidar self-supervision by occupancy estimation.
\newblock In {\em Conference on Computer Vision and Pattern Recognition (CVPR)}, 2023.

\bibitem{buchert2022exploiting}
F. Buchert, N. Navab, and S. Kim.
\newblock Exploiting diversity of unlabeled data for label-efficient semi-supervised active learning.
\newblock In {\em ICPR}, 2022.

\bibitem{nuscenes}
Holger Caesar, Varun Bankiti, Alex~H. Lang, Sourabh Vora, Venice~Erin Liong, Qiang Xu, Anush Krishnan, Yu Pan, Giancarlo Baldan, and Oscar Beijbom.
\newblock {{nuScenes}: A Multimodal Dataset for Autonomous Driving}.
\newblock In {\em Conference on Computer Vision and Pattern Recognition (CVPR)}, 2020.

\bibitem{caron2021dino}
Mathilde Caron, Hugo Touvron, Ishan Misra, Herv\'e J\'egou, Julien Mairal, Piotr Bojanowski, and Armand Joulin.
\newblock Emerging properties in self-supervised vision transformers.
\newblock In {\em International Conference on Computer Vision (ICCV)}, 2021.

\bibitem{chen2023clip2scene}
Runnan Chen, Youquan Liu, Lingdong Kong, Xinge Zhu, Yuexin Ma, Yikang Li, Yuenan Hou, Yu Qiao, and Wenping Wang.
\newblock {CLIP2Scene}: Towards label-efficient {3D} scene understanding by {CLIP}.
\newblock In {\em Conference on Computer Vision and Pattern Recognition (CVPR)}, 2023.

\bibitem{chen2021mocov3}
Xinlei Chen*, Saining Xie*, and Kaiming He.
\newblock An empirical study of training self-supervised vision transformers.
\newblock In {\em International Conference on Computer Vision (ICCV)}, 2021.

\bibitem{sspcnet}
Mingmei Cheng, Le Hui, Jin Xie, and Jian Yang.
\newblock {SSPC-Net}: Semi-supervised semantic {3D} point cloud segmentation network.
\newblock In {\em American Association for Artificial Intelligence Conference}, 2021.

\bibitem{choi2021active}
Jiwoong Choi, Ismail Elezi, Hyuk-Jae Lee, Clement Farabet, and Jose~M. Alvarez.
\newblock Active learning for deep object detection via probabilistic modeling.
\newblock In {\em International Conference on Computer Vision (ICCV)}, 2021.

\bibitem{choy2019minknet}
Christopher Choy, JunYoung Gwak, and Silvio Savarese.
\newblock {4D} spatio-temporal {ConvNets}: {Minkowski} convolutional neural networks.
\newblock In {\em Conference on Computer Vision and Pattern Recognition (CVPR)}, 2019.

\bibitem{cortinhal2020salsanext}
Tiago Cortinhal, George Tzelepis, and Eren Erdal~Aksoy.
\newblock {SalsaNext}: Fast, uncertainty-aware semantic segmentation of lidar point clouds.
\newblock In George Bebis, Zhaozheng Yin, Edward Kim, Jan Bender, Kartic Subr, Bum~Chul Kwon, Jian Zhao, Denis Kalkofen, and George Baciu, editors, {\em International Symposium on Visual Computing (ISVC)}, 2020.

\bibitem{ding2023pla}
Runyu Ding, Jihan Yang, Chuhui Xue, Wenqing Zhang, Song Bai, and Xiaojuan Qi.
\newblock {PLA}: Language-driven open-vocabulary {3D} scene understanding.
\newblock In {\em Conference on Computer Vision and Pattern Recognition (CVPR)}, 2023.

\bibitem{graham2018spconv}
B. Graham, M. Engelcke, and L. Maaten.
\newblock {3D} semantic segmentation with submanifold sparse convolutional networks.
\newblock In {\em Conference on Computer Vision and Pattern Recognition (CVPR)}, 2018.

\bibitem{distillation}
Geoffrey Hinton, Oriol Vinyals, and Jeff Dean.
\newblock Distilling the knowledge in a neural networ.
\newblock In {\em NeurIPS Deep Learning Workshop}, 2014.

\bibitem{hou2021contrastsc}
Ji Hou, Benjamin Graham, Matthias Nie{\ss}ner, and Saining Xie.
\newblock Exploring data-efficient 3d scene understanding with contrastive scene contexts.
\newblock In {\em Conference on Computer Vision and Pattern Recognition (CVPR)}, 2021.

\bibitem{hu2022sqn}
Qingyong Hu, Bo Yang, Guangchi Fang, Yulan Guo, Ales Leonardis, Niki Trigoni, and Andrew Markham.
\newblock {SQN}: Weakly-supervised semantic segmentation of large-scale {3D} point clouds.
\newblock In {\em European Conference on Computer Vision}, 2022.

\bibitem{hu2022lidal}
Zeyu Hu, Xuyang Bai, Runze Zhang, Xin Wang, Guangyuan Sun, Hongbo Fu, and Chiew-Lan Tai.
\newblock {LiDAL}: Inter-frame uncertainty based active learning for 3d lidar semantic segmentation.
\newblock In {\em European Conference on Computer Vision}, pages 248--265, 2022.

\bibitem{stocdepth}
Gao Huang, Yu Sun, Zhuang Liu, Daniel Sedra, and Kilian Weinberger.
\newblock {Deep networks with stochastic depth}.
\newblock In {\em European Conference on Computer Vision}, 2016.

\bibitem{jiang2021guided}
Li Jiang, Shaoshuai Shi, Zhuotao Tian, Xin Lai, Shu Liu, Chi-Wing Fu, and Jiaya Jia.
\newblock Guided point contrastive learning for semi-supervised point cloud semantic segmentation.
\newblock In {\em International Conference on Computer Vision (ICCV)}, 2021.

\bibitem{kong2023lasermix}
Lingdong Kong, Jiawei Ren, Liang Pan, and Ziwei Liu.
\newblock Lasermix for semi-supervised lidar semantic segmentation.
\newblock In {\em Conference on Computer Vision and Pattern Recognition (CVPR)}, pages 21705--21715, 2023.

\bibitem{lee2013pseudo}
Dong-Hyun Lee.
\newblock Pseudo-label: The simple and efficient semi-supervised learning method for deep neural networks.
\newblock In {\em ICML Workshop on challenges in representation learning (WREPL)}, 2013.

\bibitem{li2022hybridcr}
Mengtian Li, Yuan Xie, Yunhang Shen, Bo Ke, Ruizhi Qiao, Bo Ren, Shaohui Lin, and Lizhuang Ma.
\newblock {HybridCR}: Weakly-supervised 3d point cloud semantic segmentation via hybrid contrastive regularization.
\newblock In {\em Conference on Computer Vision and Pattern Recognition (CVPR)}, 2022.

\bibitem{li2022coarse3d}
Rong Li, Anh-Quan Cao, and Raoul de Charette.
\newblock {COARSE3D}: Class-prototypes for contrastive learning in weakly-supervised {3D} point cloud segmentation.
\newblock In {\em British Machine Vision Conference}, 2022.

\bibitem{lin2020segent}
Y. Lin, G. Vosselman, Y. Cao, and M.~Y. Yang.
\newblock Efficient training of semantic point cloud segmentation via active learning.
\newblock {\em ISPRS Annals of the Photogrammetry, Remote Sensing and Spatial Information Sciences}, 2:243--250, 2020.

\bibitem{liu2023weak}
Jiacheng Liu, Haiyan Guan, Xiangda Lei, and Yongtao Yu.
\newblock Weakly supervised semantic segmentation of mobile laser scanning point clouds via category balanced random annotation and deep consistency-guided self-distillation mechanism.
\newblock {\em The Photogrammetric Record}, 38(184):581--602, 2023.

\bibitem{liu2023generalized}
Kangcheng Liu, Yong-Jin Liu, Kai Tang, Ming Liu, and Baoquan Chen.
\newblock Generalized label-efficient {3D} scene parsing via hierarchical feature aligned pre-training and region-aware fine-tuning, 2023.
\newblock preprint arXiv:2312.00663.

\bibitem{liu2022ws3d}
Kangcheng Liu, Yuzhi Zhao, Qiang Nie, Zhi Gao, and Ben~M. Chen.
\newblock Weakly supervised {3D} scene segmentation with region-level boundary awareness and instance discrimination.
\newblock In {\em European Conference on Computer Vision}, 2022.

\bibitem{liu2023cpcm}
Lizhao Liu, Zhuangwei Zhuang, Shangxin Huang, Xunlong Xiao, Tianhang Xiang, Cen Chen, Jingdong Wang, and Mingkui Tan.
\newblock Cpcm: Contextual point cloud modeling for weakly-supervised point cloud semantic segmentation.
\newblock In {\em International Conference on Computer Vision (ICCV)}, 2023.

\bibitem{liu2022less}
Minghua Liu, Yin Zhou, Charles~R. Qi, Boqing Gong, Hao Su, and Dragomir Anguelov.
\newblock {{LESS}: Label-Efficient Semantic Segmentation for {LiDAR} Point Clouds}.
\newblock In {\em European Conference on Computer Vision}, 2022.

\bibitem{liu2023seal}
Youquan Liu, Lingdong Kong, Jun Cen, Runnan Chen, Wenwei Zhang, Liang Pan, Kai Chen, and Ziwei Liu.
\newblock Segment any point cloud sequences by distilling vision foundation models.
\newblock In {\em NeurIPS}, 2023.

\bibitem{liu2021otoc}
Zhengzhe Liu, Xiaojuan Qi, and Chi-Wing Fu.
\newblock One thing one click: A self-training approach for weakly supervised 3d semantic segmentation.
\newblock In {\em Conference on Computer Vision and Pattern Recognition (CVPR)}, June 2021.

\bibitem{mahmoud2023stslidr}
Anas Mahmoud, Jordan S.~K. Hu, Tianshu Kuai, Ali Harakeh, Liam Paull, and Steven~L. Waslander.
\newblock Self-supervised image-to-point distillation via semantically tolerant contrastive loss.
\newblock In {\em Conference on Computer Vision and Pattern Recognition (CVPR)}, 2023.

\bibitem{min2023occupancy}
Chen Min, Liang Xiao, Dawei Zhao, Yiming Nie, and Bin Dai.
\newblock Occupancy-{MAE}: Self-supervised pre-training large-scale lidar point clouds with masked occupancy autoencoders.
\newblock {\em IEEE Transactions on Intelligent Vehicles}, 2023.

\bibitem{segcontrast}
Lucas Nunes, Rodrigo Marcuzzi, Xieyuanli Chen, Jens Behley, and Cyrill Stachniss.
\newblock {SegContrast: 3D Point Cloud Feature Representation Learning through Self-Supervised Segment Discrimination}.
\newblock {\em IEEE Robotics and Automation Letters}, 7(2):2116--2123, 2022.

\bibitem{tarl}
Lucas Nunes, Louis Wiesmann, Rodrigo Marcuzzi, Xieyuanli Chen, Jens Behley, and Cyrill Stachniss.
\newblock Temporal consistent {3D} lidar representation learning for semantic perception in autonomous driving.
\newblock In {\em Conference on Computer Vision and Pattern Recognition (CVPR)}, 2023.

\bibitem{oquab2023dinov2}
Maxime Oquab, Timothée Darcet, Theo Moutakanni, Huy~V. Vo, Marc Szafraniec, Vasil Khalidov, Pierre Fernandez, Daniel Haziza, Francisco Massa, Alaaeldin El-Nouby, Russell Howes, Po-Yao Huang, Hu Xu, Vasu Sharma, Shang-Wen Li, Wojciech Galuba, Mike Rabbat, Mido Assran, Nicolas Ballas, Gabriel Synnaeve, Ishan Misra, Herve Jegou, Julien Mairal, Patrick Labatut, Armand Joulin, and Piotr Bojanowski.
\newblock {DINOv2}: Learning robust visual features without supervision.
\newblock {\em arXiv:2304.07193}, 2023.

\bibitem{papon2013vccs}
Jeremie Papon, Alexey Abramov, Markus Schoeler, and Florentin Wörgötter.
\newblock Voxel cloud connectivity segmentation - supervoxels for point clouds.
\newblock In {\em Conference on Computer Vision and Pattern Recognition (CVPR)}, 2013.

\bibitem{peng2023openscene}
Songyou Peng, Kyle Genova, Chiyu~"Max" Jiang, Andrea Tagliasacchi, Marc Pollefeys, and Thomas Funkhouser.
\newblock {OpenScene}: {3D} scene understanding with open vocabularies.
\newblock In {\em Conference on Computer Vision and Pattern Recognition (CVPR)}, 2023.

\bibitem{puy23waffleiron}
Gilles Puy, Alexandre Boulch, and Renaud Marlet.
\newblock Using a waffle iron for automotive point cloud semantic segmentation.
\newblock In {\em ICCV}, 2023.

\bibitem{puy2024scalr}
Gilles Puy, Spyros Gidaris, Alexandre Boulch, Oriane Siméoni, Corentin Sautier, Patrick Pérez, Andrei Bursuc, and Renaud Marlet.
\newblock Three pillars improving vision foundation model distillation for lidar.
\newblock In {\em Conference on Computer Vision and Pattern Recognition (CVPR)}, 2024.

\bibitem{radford2021clip}
Alec Radford, Jong~Wook Kim, Chris Hallacy, Aditya Ramesh, Gabriel Goh, Sandhini Agarwal, Girish Sastry, Amanda Askell, Pamela Mishkin, Jack Clark, Gretchen Krueger, and Ilya Sutskever.
\newblock {Learning Transferable Visual Models from Natural Language Supervision}.
\newblock In {\em International Conference on Machine Learning}, 2021.

\bibitem{samet2023seedal}
Nermin Samet, Oriane Siméoni, Gilles Puy, Georgy Ponimatkin, Renaud Marlet, and Vincent Lepetit.
\newblock You never get a second chance to make a good first impression: Seeding active learning for {3D} semantic segmentation.
\newblock In {\em International Conference on Computer Vision (ICCV)}, 2023.

\bibitem{sautier2024bevcontrast}
Corentin Sautier, Gilles Puy, Alexandre Boulch, Renaud Marlet, and Vincent Lepetit.
\newblock {BEVContrast}: Self-supervision in bev space for automotive lidar point clouds.
\newblock In {\em International Conference on 3D Vision (3DV)}, 2024.

\bibitem{slidr}
Corentin Sautier, Gilles Puy, Spyros Gidaris, Alexandre Boulch, Andrei Bursuc, and Renaud Marlet.
\newblock {Image-To-LiDAR Self-Supervised Distillation for Autonomous Driving Data}.
\newblock In {\em Conference on Computer Vision and Pattern Recognition (CVPR)}, 2022.

\bibitem{coreset}
Ozan Sener and Silvio Savarese.
\newblock {Active Learning for Convolutional Neural Networks: A Core-Set Approach}.
\newblock In {\em International Conference for Learning Representations (ICLR)}, 2018.

\bibitem{shi2022weakly}
Hanyu Shi, Jiacheng Wei, Ruibo Li, Fayao Liu, and Guosheng Lin.
\newblock Weakly supervised segmentation on outdoor {4D} point clouds with temporal matching and spatial graph propagation.
\newblock In {\em Conference on Computer Vision and Pattern Recognition (CVPR)}, 2022.

\bibitem{sinha20219variational}
Samarth Sinha, Sayna Ebrahimi, and Trevor Darrell.
\newblock Variational adversarial active learning.
\newblock In {\em International Conference on Computer Vision (ICCV)}, 2019.

\bibitem{su2023weakly}
Yongyi Su, Xun Xu, and Kui Jia.
\newblock Weakly supervised {3D} point cloud segmentation via multi-prototype learning.
\newblock {\em IEEE Transactions on Circuits and Systems for Video Technology}, 33(12):7723--7736, 2023.

\bibitem{spvnas}
Haotian Tang, Zhijian Liu, Shengyu Zhao, Yujun Lin, Ji Lin, Hanrui Wang, and Song Han.
\newblock Searching efficient 3d architectures with sparse point-voxel convolution.
\newblock In {\em European Conference on Computer Vision}, 2020.

\bibitem{tran2024pointct}
Anh-Thuan Tran, Hoanh-Su Le, Suk-Hwan Lee, and Ki-Ryong Kwon.
\newblock {PointCT}: Point central transformer network for weakly-supervised point cloud semantic segmentation.
\newblock In {\em IEEE Winter Conference on Applications of Computer Vision}, 2024.

\bibitem{unal2023dial}
Ozan Unal, Dengxin Dai, Ali~Tamer Unal, and Luc~Van Gool.
\newblock Discwise active learning for lidar semantic segmentation.
\newblock {\em {IEEE} Robotics and Automation Letters (RA-L)}, 8(11):7671--7678, 2023.

\bibitem{wei2023basal}
Jiarong Wei, Yancong Lin, and Holger Caesar.
\newblock {BaSAL}: Size balanced warm start active learning for lidar semantic segmentation, 2023.
\newblock preprint arXiv:2310.08035.

\bibitem{wu2021redal}
Tsung-Han Wu, Yueh-Cheng Liu, Yu-Kai Huang, Hsin-Ying Lee, Hung-Ting Su, Ping-Chia Huang, and Winston~H. Hsu.
\newblock {ReDAL: Region-Based and Diversity-Aware Active Learning for Point Cloud Semantic Segmentation}.
\newblock In {\em International Conference on Computer Vision (ICCV)}, pages 15510--15519, 2021.

\bibitem{wu2023spatiotemporal}
Yanhao Wu, Tong Zhang, Wei Ke, Sabine S{\"u}sstrunk, and Mathieu Salzmann.
\newblock Spatiotemporal self-supervised learning for point clouds in the wild.
\newblock In {\em Conference on Computer Vision and Pattern Recognition (CVPR)}, 2023.

\bibitem{xie2023annotator}
Binhui Xie, Shuang Li, Qingju Guo, Chi~Harold Liu, and Xinjing Cheng.
\newblock Annotator: An generic active learning baseline for lidar semantic segmentation.
\newblock In {\em NeurIPS}, 2023.

\bibitem{yang2022mil}
Cheng-Kun Yang, Ji-Jia Wu, Kai-Syun Chen, Yung-Yu Chuang, and Yen-Yu Lin.
\newblock An {MIL}-derived transformer for weakly supervised point cloud segmentation.
\newblock In {\em Conference on Computer Vision and Pattern Recognition (CVPR)}, 2022.

\bibitem{ye2022supervoxel}
Shanding Ye, Yongjian Fu, Hu Lin, Zhe Yin, and Zhijie Pan.
\newblock Supervoxel-based and cost-effective active learning for point cloud semantic segmentation.
\newblock In {\em ITSC}, 2022.

\bibitem{cylinder3d}
Xinge Zhu, Hui Zhou, Tai Wang, Fangzhou Hong, Yuexin Ma, Wei Li, Hongsheng Li, and Dahua Lin.
\newblock {Cylindrical and Asymmetrical 3D Convolution Networks for LiDAR Segmentation}.
\newblock In {\em Conference on Computer Vision and Pattern Recognition (CVPR)}, 2021.

\end{thebibliography}

\newpage
\clearpage
\appendix

This supplementary material is organized as follows.
 
\begin{itemize}

\item[\textbullet]  We argue about the choice of good 3D point features to build our method on (\cref{sec:good3dfeat});

\item[\textbullet]  We provide the detailed algorithm of our proposed ``Scalable SeedAL'', which is  used to select a representative subsets of scenes from SemanticKITTI~\cite{semantickitti} and nuScenes~\cite{nuscenes} (\cref{sec:sseedal});

\item[\textbullet] We present all the quantative results in a table obtained  with AL methods and our direct selection using scalable SeedAL on SemanticKITTI (\cref{sec:res_table});

\item[\textbullet] We show several visual results from our labeling pipeline (\cref{sec:visuals});

\end{itemize}

\section{The choice of good self-supervised 3D features}
\label{sec:good3dfeat}

Our method is both backbone- and pretraining-agnostic. For our experiments, we rely on recent SOTA approaches to get good 3D point features, i.e., the WaffleIron backbone \cite{puy23waffleiron} pretrained without any supervision in the ScaLR way \cite{puy2024scalr}. ScaLR can be seen as a kind of foundation model that provides 3D features for point clouds. These features are obtained by self-supervised learning, distilling 2D images features from DINOv2 \cite{oquab2023dinov2} into 3D. As shown in \cref{tab:3Dvs2D}, the corresponding 3D features lead to the best semantic segmentation performance. In particular, these image-based 3D features are superior to ``pure'' 3D features.

\begin{table*}[h!]
\caption{Resulting performance of semantic segmentation (mIoU\%) when using only 1\% of SemanticKITTI\,(SK) or nuScenes\,(NS), at different pipeline stages.}
\begin{center}
\def\tabcolsep{0.8mm}
\resizebox{0.8\linewidth}{!}{
\begin{tabular}{l|c||ccccc|ccc} 
\toprule
\multicolumn{2}{l||}{Semantic segmentation mIoU\%} & \multicolumn{5}{c|}{Image-based 3D features\raisebox{3.5mm}{}} & \multicolumn{3}{c}{Pure 3D features} \\
\midrule
&& \cite{puy2024scalr} & \cite{peng2023openscene} & \cite{liu2023seal} & \cite{caron2021dino} & \cite{chen2021mocov3} & \cite{sautier2024bevcontrast} & \cite{boulch2023also}
 & \cite{segcontrast} \\
& data & 
ScaLR & Open\raisebox{3.5mm}{} & Seal & DINO & MoCo & BEV & ALSO & Seg \\[-0.5mm]
Pipeline stage 
& set & & scene & & v1 & v3 & contrast & & contrast \\
\midrule
Frame selection \cite{samet2023seedal} \raisebox{3.5mm}{} & SK & 
& & & \textbf{52.4} & \textbf{52.4} & & 43.7 & 48.6 \\
Label propagation\,(only)
& NS & \textbf{75.5} & 68.9 &&&& 56.9 & &\\
Just finetuning \cite{puy2024scalr} & NS & \textbf{50.7} & & 45.8 & & & 37.9 & 37.4 & \\
 \bottomrule
\end{tabular}
}
\label{tab:3Dvs2D}
\end{center}
\end{table*}

Please note that DINOv2 and ScaLR require no annotation: neither tagged images nor any kind of image labeling. They are both \emph{fully un/self-supervised}.

\section{Scalable SeedAL}
\label{sec:sseedal}
This section describes how we adapted SeedAL~\cite{samet2023seedal} to make it run with thousands of samples and select diverse point clouds.

As in SeedAL, we leverage self-supervised features. But contrary to SeedAL, which uses images associated to scans and corresponding 2D features (obtained from DINOv1 \cite{caron2021dino}), we only and directly use lidar scans and 3D point features (obtained from ScaLR~\cite{puy2024scalr}, which is a distillation of DINOv2 \cite{oquab2023dinov2}). 

Similar to SeedAL, we cluster the features of each frame/scan into $k$ clusters where $k$ in the number of semantic classes. This clustering process thus yields 16 clusters for each scan in the nuScenes dataset and 19 clusters for each scan in the SemanticKITTI dataset. Even though we cannot expect a perfect clustering into the actual target classes, this provides a sensible order of magnitude to obtain a small number of clusters of features that are representative enough of the class diversity within the scene.

Subsequently, we leverage these features to assign a diversity score to a frame by averaging the dissimilarities with respect to other frames as described in Algorithm~\ref{alg: seedal}. 
The \textit{IntraSceneDiversity} procedure calculates the pairwise dissimilarities
between the cluster centers of a scene and returns the average of all these pairwise dissimilarities. Similarly, the \textit{InterSceneDiversity} procedure calculates the pairwise dissimilarities
between the cluster centers of two scenes and  returns the average of all these pairwise dissimilarities.

Those diversity computations are borrowed from SeedAL~\cite{samet2023seedal}; please refer to this paper for details. However, contrary to SeedAL that cannot scale because of the underlying linear program, whose number of variables is quadratic in the number of frames, we do not have such a limitation and, even if we still require a quadratic number of computations, we can easily scale to two orders of magnitude more frames.

\begin{algorithm}
\caption{Scalable SeedAL}
\begin{algorithmic}[1]
\State \textbf{Input:} Set of frames $F$, size limit $S$
\State \textbf{Output:} Set of diverse frames $D$

\Procedure{CalculateDiversityScores}{$F$}
    \For{each frame $f_i$ in $F$}
        \State $score_i \gets 0$
        \State  $d_{i} \gets  \text{IntraSceneDiversity}(f_i)$ 
        \For{each frame $f_j$ in $F$, $j \neq i$}
            \State $d_{j} \gets \text{IntraSceneDiversity}(f_j)$
            \State $d_{ij} \gets \text{InterSceneDiversity}(f_i, f_j)$
            \State $score_i \gets score_i + d_{i}*d_{j}*d_{ij}$
        \EndFor
        \State $score_i \gets \frac{score_i}{|F| - 1}$
    \EndFor
    \State \textbf{return} $\{score_1, score_2, \ldots, score_{|F|}\}$
\EndProcedure

\Procedure{SelectDiverseFrames}{$F$, $S$}
    \State $scores \gets \Call{CalculateDiversityScores}{F}$
    \State Sort $F$ by $scores$ in descending order
    \State $D \gets$ first $S$ frames of $F$
    \State \textbf{return} $D$
\EndProcedure
\end{algorithmic}
\label{alg: seedal}

\end{algorithm}
\vspace{-3mm}
\section{Quantitative results}
\label{sec:res_table}

To make it easier to quantitatively compare performances in Figure 3 of the main paper, we report here in Table~\ref{tab:results} the detailed quantitative results obtained from AL methods with random and SeedAL initilization, and direct selection using scalable SeedAL on SemanticKITTI~\cite{semantickitti} dataset. 

\begin{table*}
    \centering
    \caption{Performance (\% mIoU)  comparison of our direct (non iterative) data selection strategy MILAN with SOTA active learning strategies on SemanticKitti~\cite{semantickitti}. All models are trained using SPVCNN as network architecture and using solely the scans with labeled points for training (no semi-supervision). }
    \label{tab:results}
    \setlength{\tabcolsep}{5pt}
    \resizebox{0.5\linewidth}{!}{
    \begin{tabular}{l c c c c}
     \toprule
    Method & 1\% & 2\% & 3\% & 4\%   \\ 
    \midrule
     \textit{Iterative Selection} & & & &    \\ 
    CoreSet~\cite{coreset} & 46.4 & 49.5 & 52.1 &  54.1 \\ 
    CoreSet~\cite{coreset}+SeedAL~\cite{samet2023seedal} & 52.1 & 53.8 &  55.2 & 56.9 \\
    \midrule
    ReDAL~\cite{wu2021redal} & 46.5 & 53.8 &  56.57 &  58.4   \\
    ReDAL~\cite{wu2021redal}+SeedAL~\cite{samet2023seedal} & 50.5 & 53.9 & 55.8 &   58.9  \\
    \midrule
    LiDAL~\cite{hu2022lidal}       & 48.8 & 57.1 &  58.7 &  59.3   \\
    LiDAL~\cite{hu2022lidal} +SeedAL~\cite{samet2023seedal} & 52.6 & 57.8 & 59.3 &   60.3  \\
    \midrule
    \textit{Direct Selection } & & & &    \\ 
    Scalable SeedAL & 52.3 &  57.4 & 58.2 &   60.6  \\ 
    \midrule
    \end{tabular} 
    }    

\end{table*}

\section{Qualitative analysis}
\label{sec:visuals}

\begin{figure*}
    \centering
    \begin{minipage}{0.49\linewidth}
    \centering\scriptsize Our propagated labels

    \end{minipage}
    \begin{minipage}{0.49\linewidth}
    \centering\scriptsize Ground truth labels

    \end{minipage}
    \begin{minipage}{0.49\linewidth}
    \includegraphics[width=\linewidth]{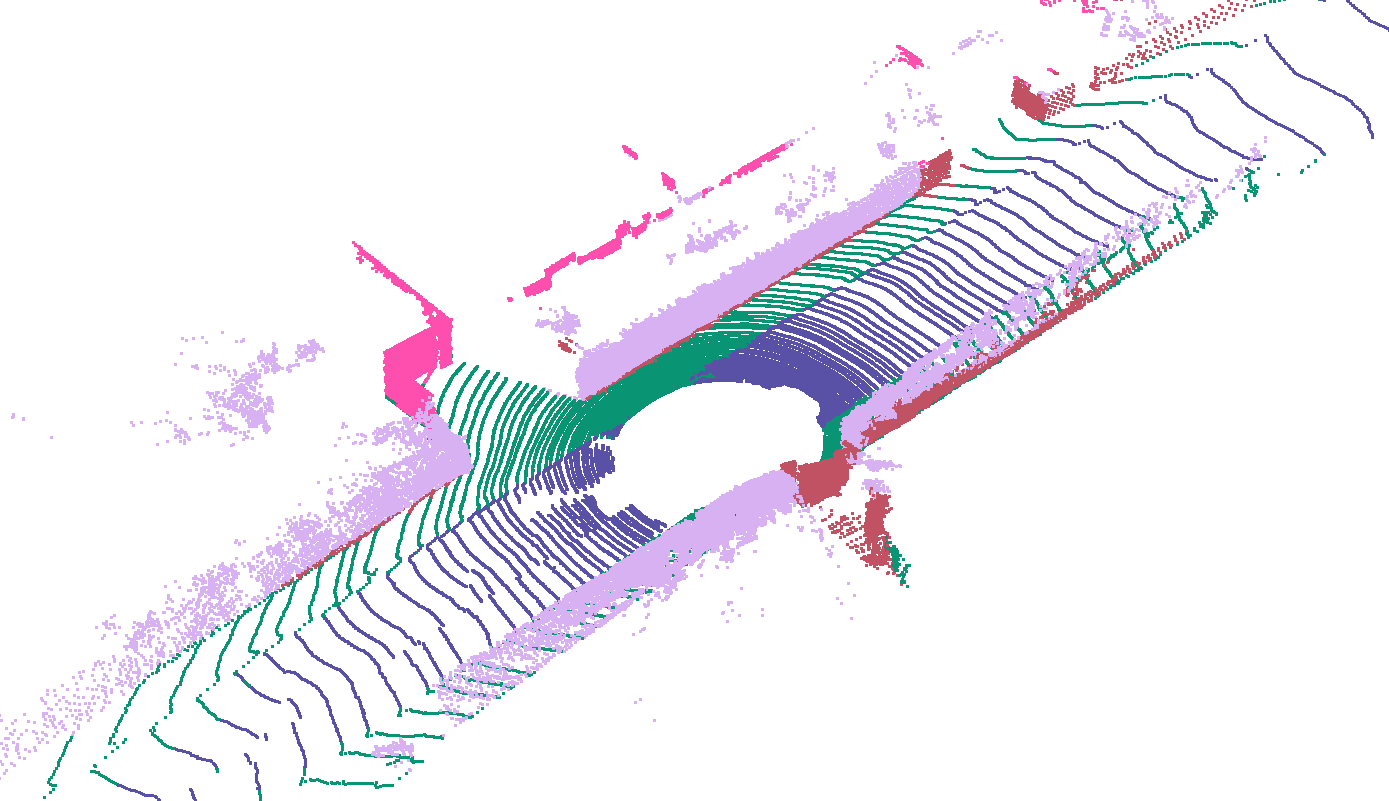}    
    \end{minipage}
    \begin{minipage}{0.49\linewidth}
    \includegraphics[width=\linewidth]{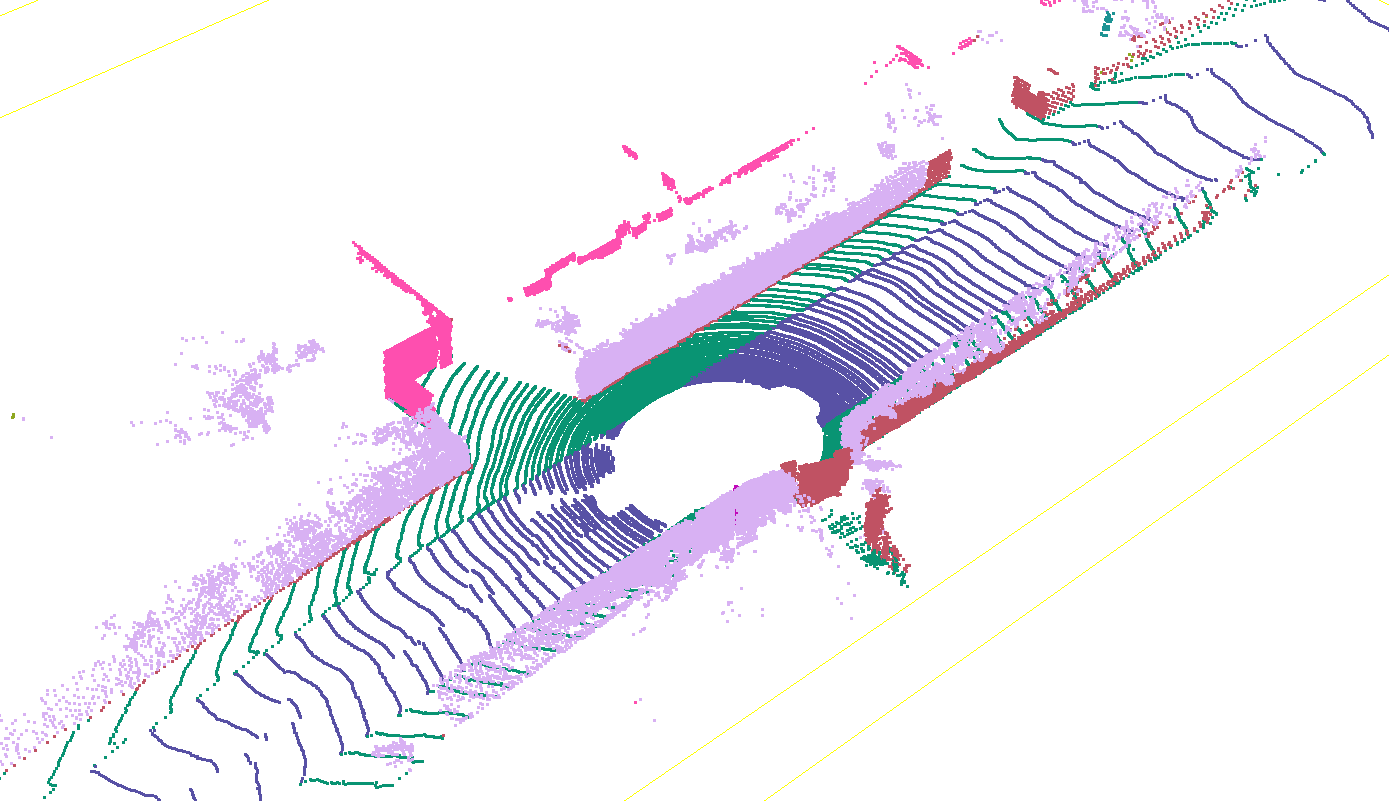}    
    \end{minipage}

    \begin{minipage}{0.49\linewidth}
    \includegraphics[width=\linewidth]{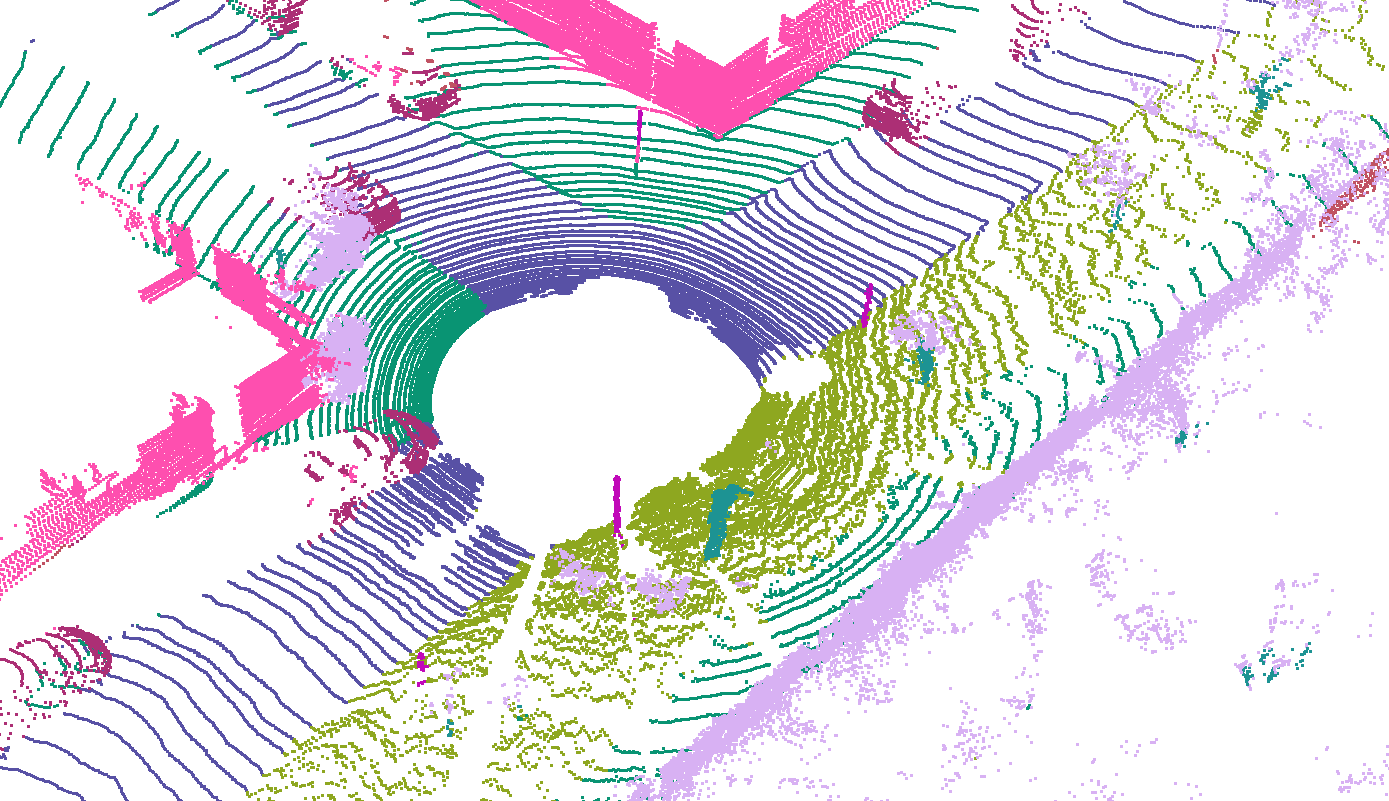}    
    \end{minipage}
    \begin{minipage}{0.49\linewidth}
    \includegraphics[width=\linewidth]{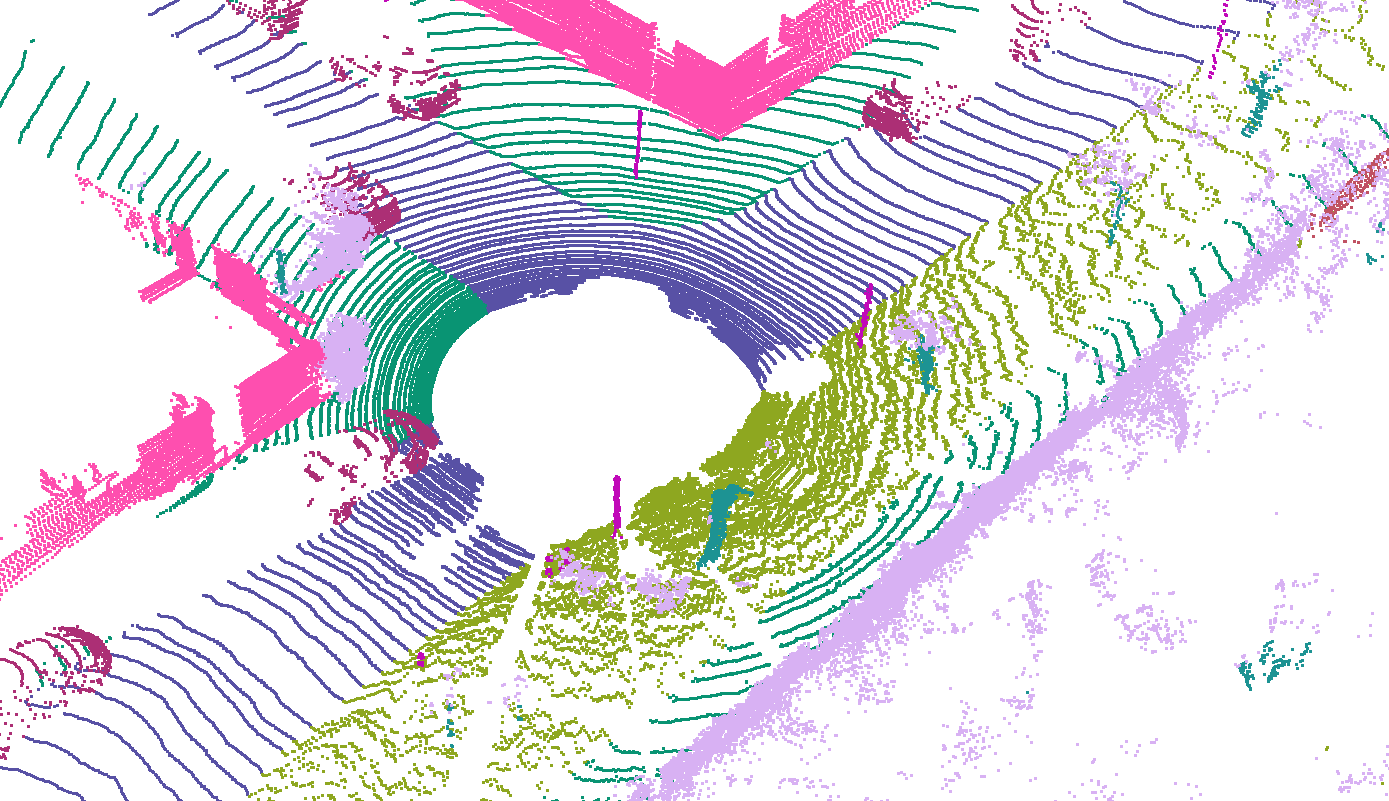}    
    \end{minipage}
    \begin{minipage}{0.49\linewidth}
    \includegraphics[width=\linewidth]{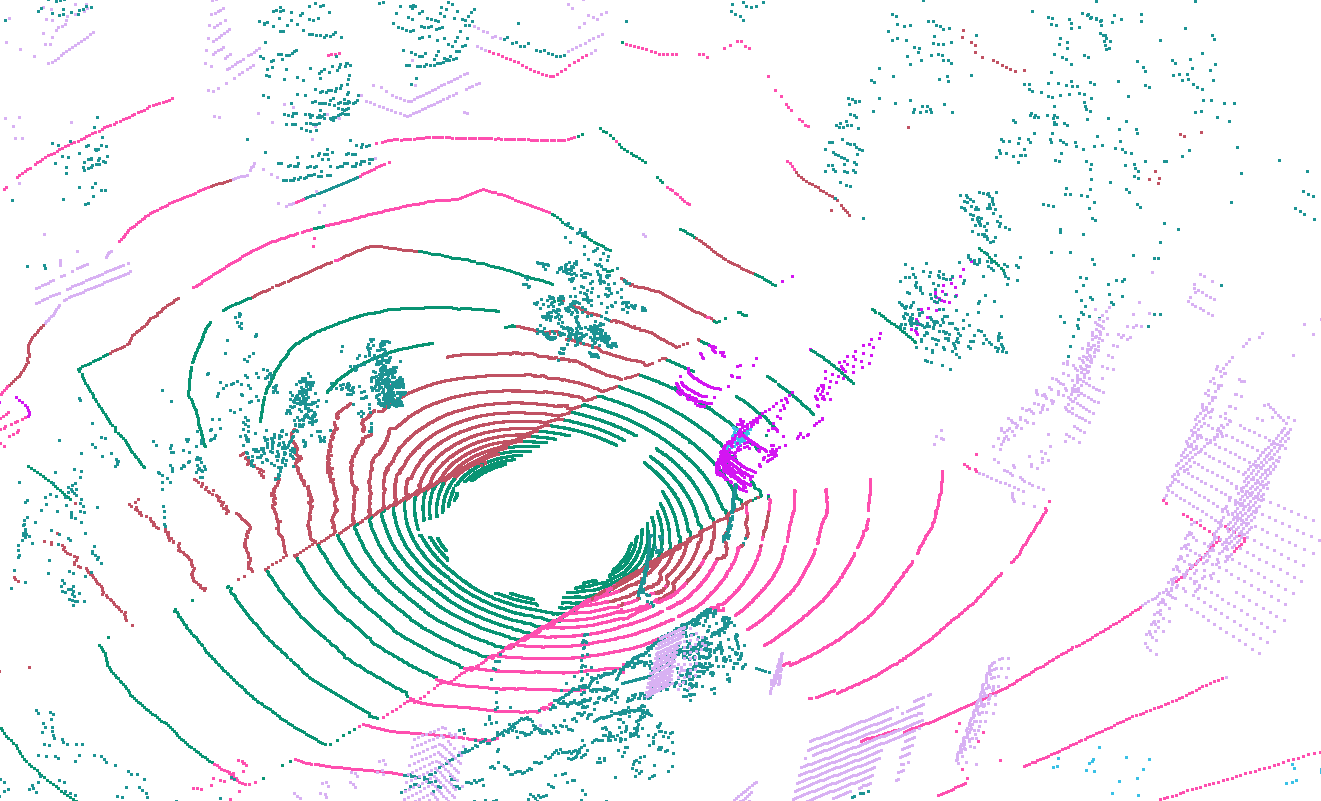}    
    \end{minipage}
    \begin{minipage}{0.49\linewidth}
    \includegraphics[width=\linewidth]{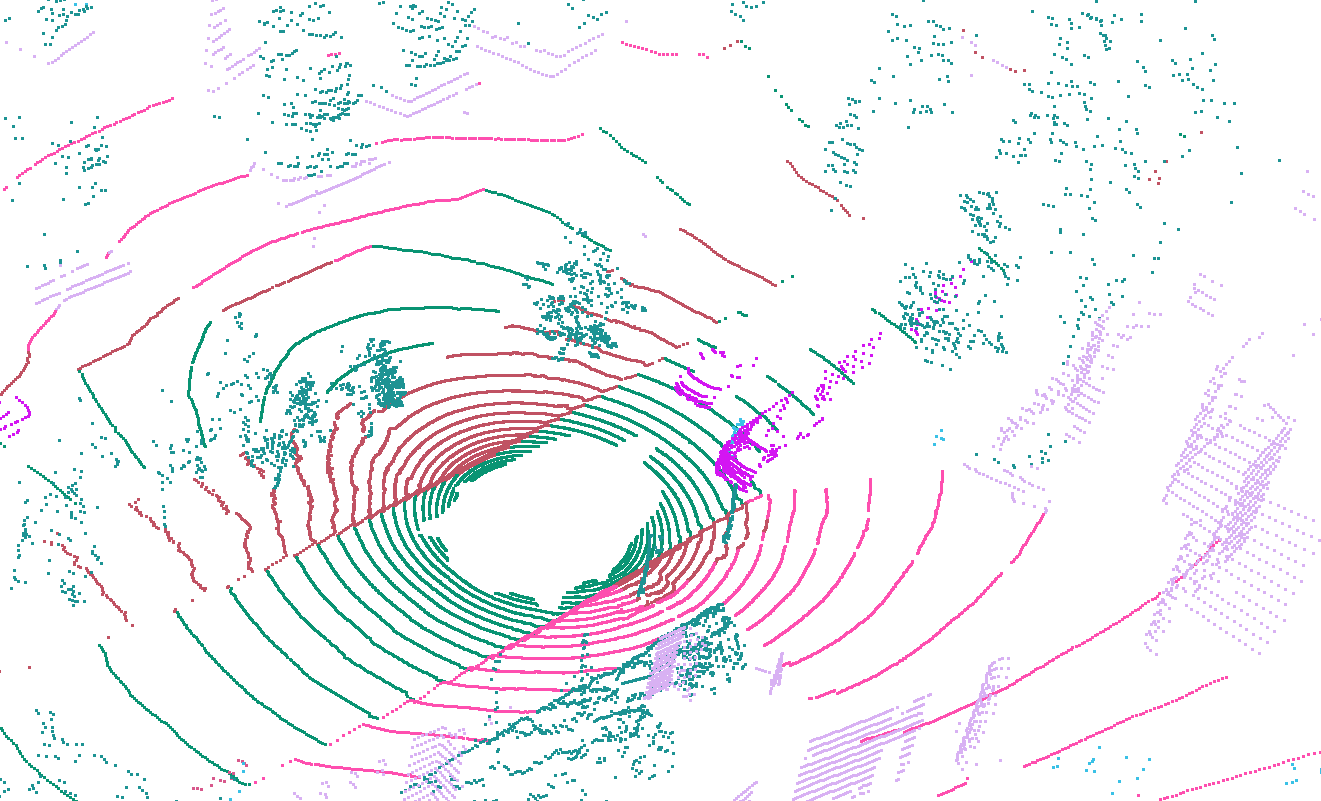}    
    \end{minipage}
    \begin{minipage}{0.49\linewidth}
    \includegraphics[width=\linewidth]{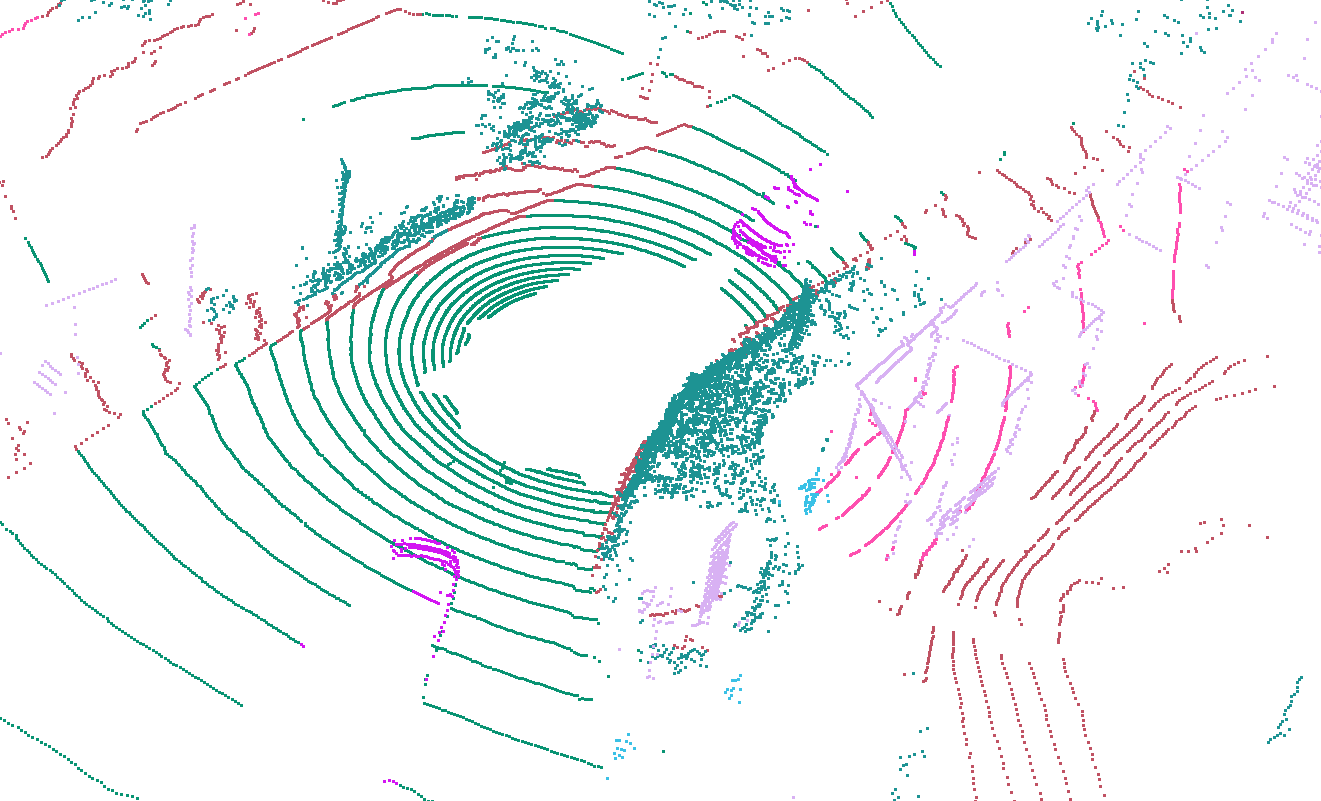}    
    \end{minipage}
    \begin{minipage}{0.49\linewidth}
    \includegraphics[width=\linewidth]{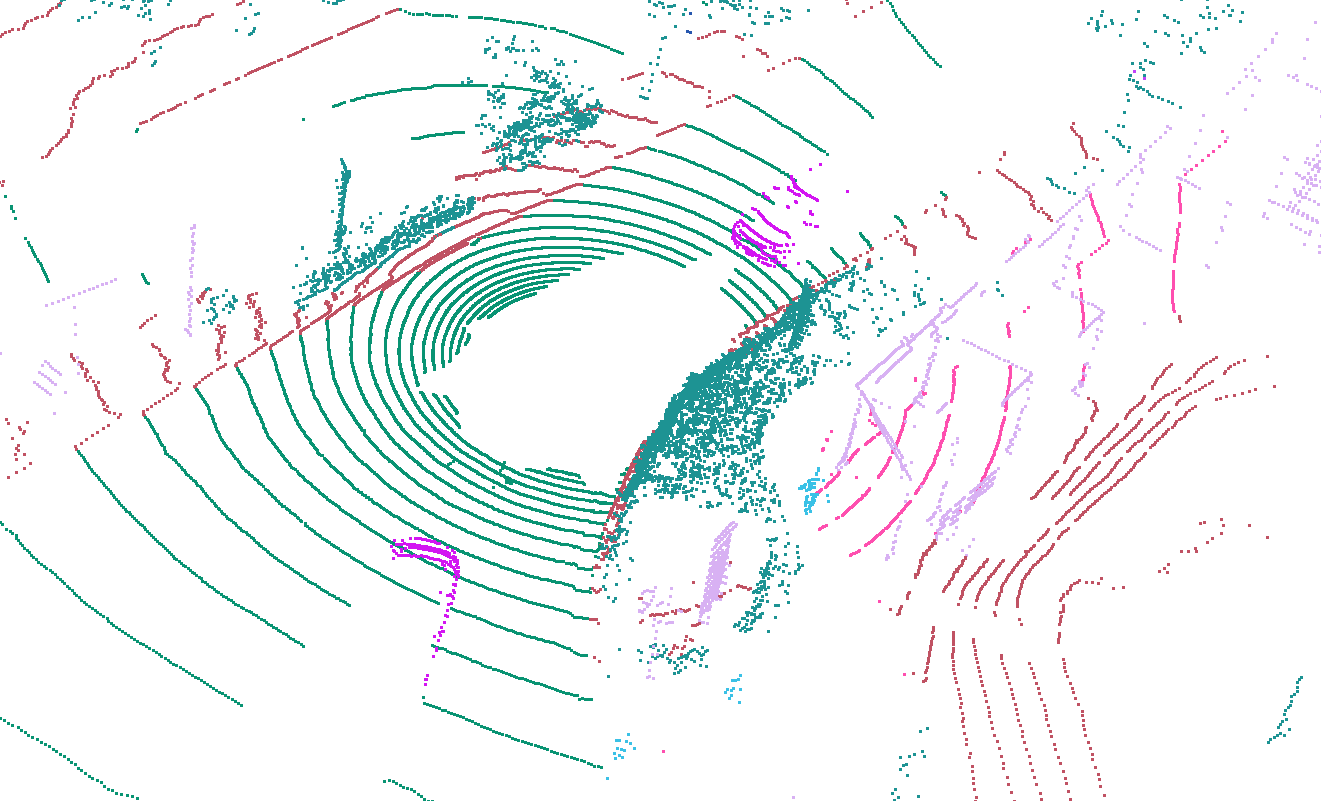}    
    \end{minipage}
    \caption{\textbf{Quality of our propagated labels}. Left: Our propagated labels after annotating 1\% and 2\% of the points per scan on SemanticKITTI (first two rows) and nuScenes (last two rows), respectively. Right: Ground truth labels.}
    \label{fig:visual_quality}
\end{figure*}

We present in \cref{fig:visual_quality} examples of annotated scans using our method. These annotations can be compared to the corresponding ground truth. It confirms visually the numbers presented in Sec.~\textcolor{red}{4.3}: in all cases our label propagation method leads to pseudo-labels of high quality.

\end{document}